\pdfoutput=1
\documentclass{article} 
\usepackage{iclr2027_conference_arXiv,times}


\usepackage{amsmath,amsfonts,bm}

\def\eqref#1{equation~\ref{#1}}

\def\1{\bm{1}}

\DeclareMathAlphabet{\mathsfit}{\encodingdefault}{\sfdefault}{m}{sl}
\SetMathAlphabet{\mathsfit}{bold}{\encodingdefault}{\sfdefault}{bx}{n}

\usepackage{hyperref}
\usepackage{url}

\usepackage{booktabs}
\usepackage{amsfonts,amsmath,amssymb}
\usepackage{graphicx}
\usepackage{natbib}
\usepackage{wrapfig}

\usepackage{longtable}
\usepackage{multirow}
\usepackage[table]{xcolor}
\usepackage{listings}
\usepackage{amssymb}
\usepackage[framemethod=TikZ]{mdframed}
\usepackage[section]{placeins}
\iclrfinalcopy
\mdfdefinestyle{participantbox}{
  backgroundcolor=gray!10,
  linecolor=gray!40,
  linewidth=0.5pt,
  roundcorner=2pt,
  innertopmargin=6pt,
  innerbottommargin=6pt,
  innerleftmargin=8pt,
  innerrightmargin=8pt,
  skipabove=6pt,
  skipbelow=6pt,
}
\newmdenv[style=participantbox]{participanttextbase}
\newenvironment{participanttext}
  {\begin{participanttextbase}\setlength{\parindent}{0pt}\setlength{\parskip}{6pt}}
  {\end{participanttextbase}}

\begin{document}

\title{How Much Do LLM-as-a-Judge Design Choices Matter? A Systematic Comparison of Prompt Designs, Rating Scales, and Models}
\author{
  \textbf{Laurène Vaugrante}\thanks{Corresponding author: \texttt{laurene.vaugrante@iris.uni-stuttgart.de}}\hspace{0.5em},\hspace{1em}
  \textbf{Thilo Hagendorff} \\
  AI Safety Research Group, University of Stuttgart
}


\maketitle

\begin{abstract}
Researchers increasingly use Large Language Models as judges (LLM-as-a-judge) to evaluate model outputs. Yet there are no standards for how to design these judges. Typically, researchers choose the prompt, rating scale, and model intuitively. If these choices change the judge's verdicts, two studies can reach different conclusions about the same facts. To address this risk and to provide an empirical basis for judge designs, we evaluate 10 reasoning models across multiple designs on two tasks: a scalar rating of sentence sentiment and toxicity (over 500 items per category), as well as a binary accuracy classification of question-answer pairs ($n=600$). For the rating tasks, despite judges showing significant disagreements with the human ground truth, the practical size of differences is small enough to consider most judges reliable (mean absolute deviation of 0.11 points on a 1 -- 7 scale); toxicity judges even outperform standard classifiers. Judges are also highly accurate on average (96.5\%) for the accuracy classification task. However, design choices can produce shifts: changing the rating scale alone can shift measured bias by up to 0.93 points (rating task), and while accuracy levels are rarely impacted, design choices consistently impact judge leniency (classification task; leniency drop of 28.9 percentage points when using detailed prompts, and up to 56.1 percentage points when switching models). Counterintuitively, lower reasoning effort affects neither accuracy nor leniency. Across both tasks, model identity is the dominant source of variance.
These findings suggest that while LLM judges are broadly trustworthy in aggregate, design choices can be meaningful sources of variance. Given the growing reliance on automated evaluation in LLM research, we intend this study as a methodological reference for designing more robust and replicable LLM-as-a-judge pipelines.

\end{abstract}


\section{Introduction}

The broad adoption of Large Language Models (LLMs) across research, industry, and public-facing applications has made the evaluation of their outputs necessary \citep{chang2024surveyevaluation, minaee2024largelanguagemodelssurvey}. Many tasks on which LLM judges are deployed, such as open-ended dialogue, summarization, or complex reasoning, require nuanced, subjective judgment, and semantic understanding \citep{guzman2015humans, gillick2010nonexpert}. This judgment is costly to obtain from human annotators at scale and poorly captured by traditional metrics such as BLEU \citep{papineni2002bleu} or ROUGE \citep{lin2004rouge}. This motivates the use of LLMs themselves as automated judges \citep{chiang2023alternative}. \citet{zheng2023judging} established that a strong LLM judge can match human-level agreement on open-ended chat quality assessment, and follow-up work extended this validation across other Natural Language Generation (NLG) tasks such as summarization, translation, and general scoring \citep{fu2024gptscore, wang2023chatgptnlgevaluator}. This validation and adoption of LLM judges in NLG \citep{liu2023geval,fu2023factuality,tong2024codejudge} seems to largely overshadow human annotation: a recent analysis of NLG papers found that fewer than 8\% of studies employing an LLM judge explicitly validate it against human judgment, and that automatic metrics, LLM judges, and human evaluators are applied to different tasks and criteria rather than compared directly \citep{yang2026measuring}. Meanwhile, there remains no consensus on standardized evaluation protocols, and systematic comparisons of LLM judge prompt designs remain rare \citep{li2025robustjudge, li2026scoringbias}. Furthermore, LLM judges have been shown to exhibit systematic biases, including position bias \citep{zheng2023judging}, verbosity bias \citep{wu2025styleoversubstance}, and self-preference bias \citep{wataoka2024selfpreference}. Research has also repeatedly shown that LLMs are highly sensitive to prompt formulation \citep{mizrahi2024stateofwhatart, sun2024zeroshotrobustness, sclar2024quantifying, chatterjee2024posix, gu2022taskinstructionrobustness, cao2024worstprompt}, with prompt templates sometimes exerting a greater influence on outputs than the underlying question itself \citep{liu2026promptsensitivity}. Given the risk that design choices substantially affect the resulting annotations and, by extension, the conclusions drawn from them, investigating the sensitivity of LLM judges is critical.

\begin{figure}[h]
    \begin{center}
    \includegraphics[width=\linewidth]{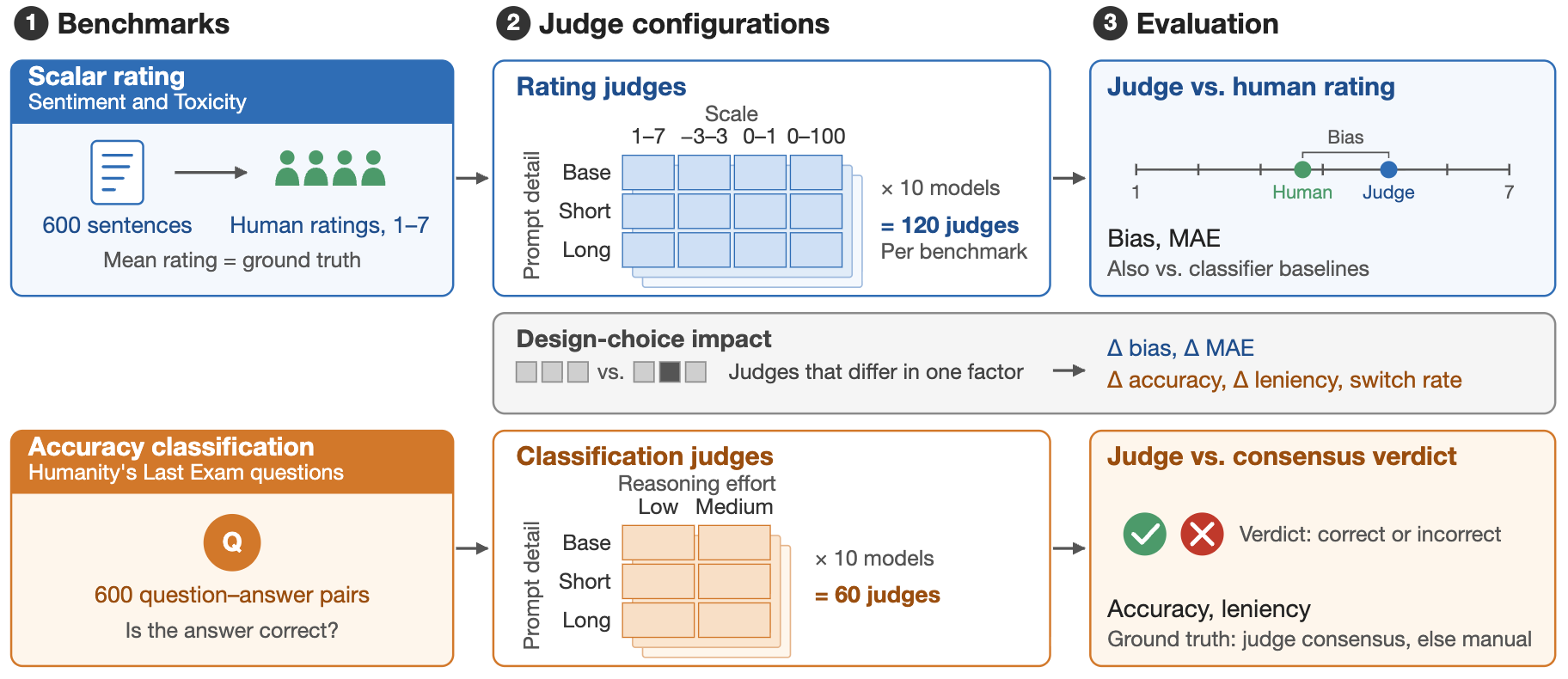}
    \end{center}
    \caption{Overview of our experiments.}
\label{fig:illustration}
\end{figure}

In this study, we examine this across two qualitatively different evaluation setups: subjective scalar ratings (sentence toxicity and sentiment) and binary response classifications (accuracy of answers). For the scalar tasks, we collected human annotations as ground truth against which LLM ratings are compared. We measure the performance of a large set of LLM judges, varying their design, and assessing whether the judges' design choices impact their performance (Figure \ref{fig:illustration}). 
Our results reveal that, on average, LLM judges are reliable regardless of their design: most design choices produce only small differences in performance, and this holds even for counterintuitive choices such as reducing reasoning effort. This reliability, however, does not imply a uniform behavior: judges reaching comparable accuracy can showcase drastically different leniency rates. Model identity is the dominant source of variance across both tasks, and no single model is consistently the best. Robustness also degrades, and design choices matter more, as task subjectivity increases (toxicity assessments, ambiguous binary response classifications). These findings offer concrete guidance for deploying and configuring LLM judges reliably.

\section{Related Work}
\label{sec:related_work}

The LLM-as-a-judge method has been mapped by several surveys \citep{gu2024llmjudgesurvey, li2024llmsasjudges}, which found that while LLM judges can be reliable evaluators for some tasks, such as instruction-following and mathematical reasoning-trace verification, they perform worse on other tasks, notably toxicity and safety judgments \citep{bavaresco2025llms}. Research is similarly divided on the inter-annotator agreement scores. Some studies find that LLM judges remain behind inter-human agreement, particularly when judge prompts are complex or lengthy \citep{thakur2024judgingjudges}; others report the opposite: \citet{zheng2023judging} notably found that while a GPT-4 judge matched the human inter-annotator agreement for writing and roleplay tasks, this agreement rate lowered on categories such as math and reasoning, where models made verifiable errors. In addition to the LLM judge biases mentioned in the introduction \citep{zheng2023judging, wu2025styleoversubstance, wataoka2024selfpreference}, they have also been shown to rely on superficial cues rather than genuine response quality \citep{marioriyad2025silentjudge}. \citet{verga2024replacing} show that replacing a single judge with a panel of smaller, more diverse judges improves correlation with human judgment, reduces self-preference bias, and costs substantially less, which confirms that the LLM judge design can impact its reliability. Furthermore, the design choices underlying LLM-as-a-judge methodologies have been shown to affect judge reliability. \citet{li2025robustjudge} and \citet{wei2024systematic} show that the choice of prompt design and of evaluator model substantially affect judge robustness. Complex or lengthy prompts have also been shown to reduce judge accuracy \citep{thakur2024judgingjudges}, while reference-guided grading has been shown to reduce judge failure specifically on math and reasoning tasks \citep{zheng2023judging}. \citet{gu2024llmjudgesurvey} also note that most prior bias research has focused on comparing which response out of two given answers was the best, rather than scalar scoring, leaving the reliability of scalar LLM-judge ratings underexplored.

Beyond prompt content, formatting constraints placed on a model's output affect performance. \citet{do2024formatbias} find that output formats significantly affect performance even when the underlying task is unchanged. \citet{tam2024letmespeak} similarly show that structured-output constraints, such as making the model output a JSON object, impact task performance, although the direction and magnitude of the effect are task-dependent. Such constraints tend to degrade performance on open-ended reasoning tasks, while performance on classification tasks is unaffected or improved. Similarly, the methodology used to extract answers from reasoning models' outputs can change which models perform best \citep{jo2025findinganswers}, which confirms that, beyond the evaluation method itself, output-level methodological decisions can also shape the results; similar effects may hold for LLMs deployed as judges.
Within LLM-judge scoring specifically, scoring granularity and range both introduce bias \citep{stureborg2024inconsistent, li2026gradingscale}, with diverging results regarding the effect of scale choice on judge reliability. More broadly, similar concerns extend to LLM safety evaluations: \citet{beyer2025llmsafety} show that inconsistent evaluator choices alone can make results incomparable across papers, echoing concerns about evaluation reliability \citep{vaugrante2025replicability, gu2024llmjudgesurvey}. 

Taken together, these mixed findings establish that while LLM-as-a-judge methods are widely adopted, they are prone to bias, and that the specific design choices underlying them, such as the level of prompt detail or the model used, are rarely varied and compared within a single study. Our work addresses this gap by varying prompt designs and scale formats across 10 model judges. Moreover, we compare these results to human annotations as ground truth, and broadly used classifiers for the scalar rating tasks.

\section{Methods}
\label{sec:methods}

\subsection{Benchmarks and Data}

We evaluate LLM-as-a-judge designs on outputs of two tasks: scalar rating, covering sentiment and toxicity, as well as binary accuracy classification. All benchmarks are publicly available \footnote{\url{https://osf.io/sra9e/}}.

\textbf{Scalar Rating Benchmarks (Sentiment and Toxicity).} We constructed both benchmarks by first generating $n=600$ items per dimension using Grok 4.1~\citep{xai2025grok41}, with varying levels of toxicity or sentiment (see the generation prompt in Appendix~\ref{app:benchmark_gen}). To establish each sentence's toxicity or sentiment ground truth score, we aggregate ratings from human annotators. The 600 benchmark items were divided into a split of sentences used exclusively for the evaluation (501 sentiment, 521 toxicity), and a support split for methods requiring in-context examples (99 sentiment, 79 toxicity). When selecting support examples, we prioritized items where there was a general consensus on the score (high inter-annotator agreement), and whose mean scores fell near the anchors of the scale, so as to provide unambiguous anchor examples for few-shot in-context-learning prompting. For the maximum anchor on the 1--7 scale, any sentence with a mean annotator score of $\geq 6.7$ was treated as a scale-maximum example, given that the theoretical maximum of 7.0 was not present in the annotated scores.

\textbf{Binary Accuracy Classification Benchmark.} Question-answer pairs ($n=600$) were sourced from Humanity's Last Exam (HLE) \citep{phan2025hle} to include high-difficulty questions on diverse topics. They were filtered to exclude items requiring supplementary information such as images, and to exclude multiple-choice questions. The candidate responses to classify by our LLM judges were generated by Gemini 3.1 Pro~\citep{deepmind2026gemini31pro} using a sampling temperature of 0.7.

\textbf{Human Annotation.} For rating benchmarks, surveys were administered via LimeSurvey, with participants recruited through Prolific using a representative sample and compensated at an average rate of £9.95/hour. Each item from the benchmarks was rated by several annotators ($N_{\text{sentiment}} = 10$, $N_{\text{toxicity}} = 15$) on a 1--7 Likert scale. Full annotator instructions are provided in Appendix~\ref{app:annotating_instructions}. Consistent with prior literature on subjective annotation tasks \citep{artstein2008survey, vidgen2020directions}, inter-annotator agreement was relatively high for both dimensions (Krippendorff's $\alpha_{\text{sentiment}}$ = 0.86; Krippendorff's $\alpha_{\text{toxicity}}$ = 0.68), with a lower agreement reached for the Toxicity task due to its highly subjective nature. This study received ethics board approval for conducting these surveys.

\subsection{LLM-as-a-Judge designs}

\textbf{Scale.} For the scalar rating tasks, we vary the numerical range of the requested LLM judge output across four formats: a \textbf{1 -- 7 integer scale}, a \textbf{-3 -- 3 integer scale}, a \textbf{0 -- 1 decimal scale}, and a \textbf{0 -- 100 integer scale} (see Appendix~\ref{app:prompts}). For each scale, the scoring precision expected is implied by the instructions and examples in the Short In-Context prompts and Long In-Context prompts, where integers are shown for every scale except 0 -- 1, which shows two decimal places (see Appendix \ref{app:prompts}). 

\textbf{Prompt Detail.} We define three levels of prompt detail. The \textbf{Base Prompt} provides a minimal task instruction and the response scale when necessary (rating tasks). The \textbf{Short In-Context Prompt} expands this with more detailed instructions and examples ($n=3$ for rating tasks, $n=4$ for the classification task). The \textbf{Long In-Context Prompt} further extends this with detailed scoring instructions for each value anchor of the given response scale and explicit scoring principles (rating tasks) or detailed classification instructions and principles (classification task), a step-by-step reasoning procedure, and a larger set of examples. Prompt examples are provided in Appendix \ref{app:prompts}. 

\textbf{Models.} We evaluate \textbf{10 reasoning models} spanning a range of capability tiers and providers: gpt-oss 120b~\citep{gptoss2025}, GPT-5.5~\citep{openai2026gpt55}, GPT-5.6 Terra~\citep{openai2026gpt56terra}, Gemini 3.7 Flash~\citep{deepmind2026gemini37flash}, DeepSeek V4 Pro~\citep{deepseekv42026}, Claude Haiku 4.5~\citep{anthropic2025haiku45}, Claude Sonnet 5~\citep{anthropic2026sonnet5}, Kimi K3~\citep{kimik32026}, Qwen 3.8
27B~\citep{qwen32025}, and GLM-5.3~\citep{glm5team2026}.

\textbf{Reasoning Effort.} For the classification task, we vary each model's reasoning effort parameter, which controls the amount of internal reasoning a model performs before producing its verdict. We compared models with a \textbf{low reasoning effort} and a \textbf{medium reasoning effort} (default setting).

\subsection{Evaluation Methodology}
\subsubsection{Scalar Rating} 

For both benchmarks, we construct all 120 possible judge designs by crossing each design choice: prompt detail, rating scale, and model. Each design rates every benchmark item, returning its score as a JSON object; we then parse this output and rescale the extracted score to the 1--7 range to obtain our judge assessments. We compare these assessments with the human-annotator ground truth.
To contextualize our comparisons, we also use popular classifiers on each benchmark item and measure their performance compared to the established ground truth. For Sentiment, we use VADER \citep{hutto2014vader} and TextBlob \citep{textblob}, which are lexicon- and rule-based sentiment classifiers (sentiment score on a -1 -- 1 scale); for Toxicity, we use Detoxify \citep{hanu2020detoxify}, a BERT-based classifier trained on the Jigsaw Toxic Comment Classification challenges, and Perspective API \citep{lees2022perspective}, Google Jigsaw's toxicity classifier, trained on crowd-sourced human ratings of perceived toxicity (toxicity probability on a 0 -- 1 scale). Classifier assessments were also rescaled onto the 1 -- 7 range for comparison purposes.

\textbf{Metrics.} We measure the \textbf{bias} as the signed mean difference between the normalized LLM judge scores and the human ground truth, and record its sign as \textbf{direction} to capture the overall tendency and control for systematic over- or under-rating. However, a low bias does not necessarily imply low per-item difference, as positive and negative values can cancel out in a signed average: we therefore use the \textbf{mean absolute error (MAE)} to estimate the mean per-item difference. All bias and MAE values reported are relative to the 1 -- 7 scale. 
We note the \textbf{rank} of each LLM judge across all prompt designs, using a high Krippendorff's $\alpha$ against human ground truth as quality indicator. While this is useful to identify which setting performs best, it does not account for the magnitude: a low rank can be associated with a negligible average bias and vice versa, depending on the per-item variability or performance distribution across judges. To capture how much a design choice can reorder this ranking, we additionally report \textbf{Best} as the
average change in rank position observed when that design choice alone is varied.

\textbf{Statistical Significance.} Given the size of our experiments where each per-item comparison draws on several hundred observations, small differences that would be negligible in practice frequently reach statistical significance. Therefore, we do not consider this to be a reliable indicator of effect magnitude in our setting and instead report the effect strength using a rank-biserial correlation (RBC). We nevertheless calculate and seldom mention significance values, notably in the Appendix tables, using a paired Wilcoxon signed-rank test with a Benjamini-Hochberg correction applied to account for the number of comparison groups. When significance results are presented in tables, we use the following representation: $^{*}p<.05$, $^{**}p<.01$, $^{***}p<.001$, ns = not significant.

\subsubsection{Accuracy Classification} 

We construct all 60 possible judge designs by crossing prompt detail, reasoning effort, and model. For each benchmark item, every judge is given the question, the LLM-generated answer to classify, and the verified ground-truth answer, and is asked to return a binary correct/incorrect verdict as a JSON object. Case-insensitive variants of ``true'' and ``false'' are treated as equivalent during parsing. Reasoning traces are stored when available. To establish the classification ground truth, we partly use the evaluated judges' own classification outcomes: an item's accuracy label is determined by majority vote whenever at least 90\% of the 60 designs agree, and the remaining items are resolved manually. This yields three subsets: 429 \textit{full-agreement} items (unanimous verdict across all 60 designs), 124 \textit{near-agreement} items (agreement rate $\geq 90\%$ and $< 100\%$), and 47 \textit{ambiguous} items (agreement rate $< 90\%$, manually resolved).

\textbf{Metrics.} To analyze the obtained results, we measure \textbf{accuracy} as the proportion of question-answer pairs correctly classified against the established ground truth, and \textbf{leniency} as the proportion of a judge's errors that are false positives (incorrectly judging a response as correct) rather than false negatives, to capture any systematic over- or under-forgiving tendencies. However, similar accuracy or leniency rates can still hide different underlying behavior on a per-item basis: we therefore additionally report the \textbf{switch rate} as the proportion of items whose verdict changes when a single design choice is varied, to capture instability.

\textbf{Statistical Significance.} As with the rating tasks, we do not treat statistical significance alone as a reliable indicator of effect magnitude and report our selected metrics alongside any significance statistic, namely chi-square test of independence for accuracy and leniency impact, and Cochran's Q test for switch rate impact, both Benjamini-Hochberg corrected within each item-difficulty tier. When significance results are presented in tables, we use the same significance representation as in the rating tasks. 

\section{Results}

\subsection{Rating Benchmarks}
\label{sec:results-rating}
To interpret our results, we focus on two overarching questions. First, to assess whether LLM judges are a suitable substitute for human annotation or non-LLM classifiers, we examine \textbf{how LLM judges' and classifiers' scores align with the human ground truth} (Figure \ref{fig:design-sensitivity}, Table \ref{tab:ground-truth-significance}). Second, we examine \textbf{how much LLM judges' scores vary depending on each design choice} to determine whether these choices can artificially skew assessments (Table \ref{tab:star-summary}). As Sentiment and Toxicity judges behave differently across most patterns, we analyze both benchmarks separately throughout this section. 
\subsubsection{Do the LLM judges disagree with humans and classifiers?}
\label{sec:rating-diff_groundtruth}
\textbf{LLM judges reliably rate the sentiment and toxicity levels of sentences.}
Among the 240 designs evaluated across both benchmarks, 67\% of LLM judges obtain assessments that significantly differ from the human ground truth. However, the observed effect remains small: only 24\% of designs show a medium to large effect strength, and 7\% deviate by more than 0.3 points on the 1 -- 7 scale. Overall, the practical implications remain modest, with judges deviating by only 0.11 points on average (values ranging from 0.00 to 0.76 across all designs; see Table \ref{tab:q1-avg-appendix}, Appendix \ref{app:extended_results_rating}). 

\begin{figure}[h]
    \begin{center}
    \includegraphics[width=0.8\linewidth]{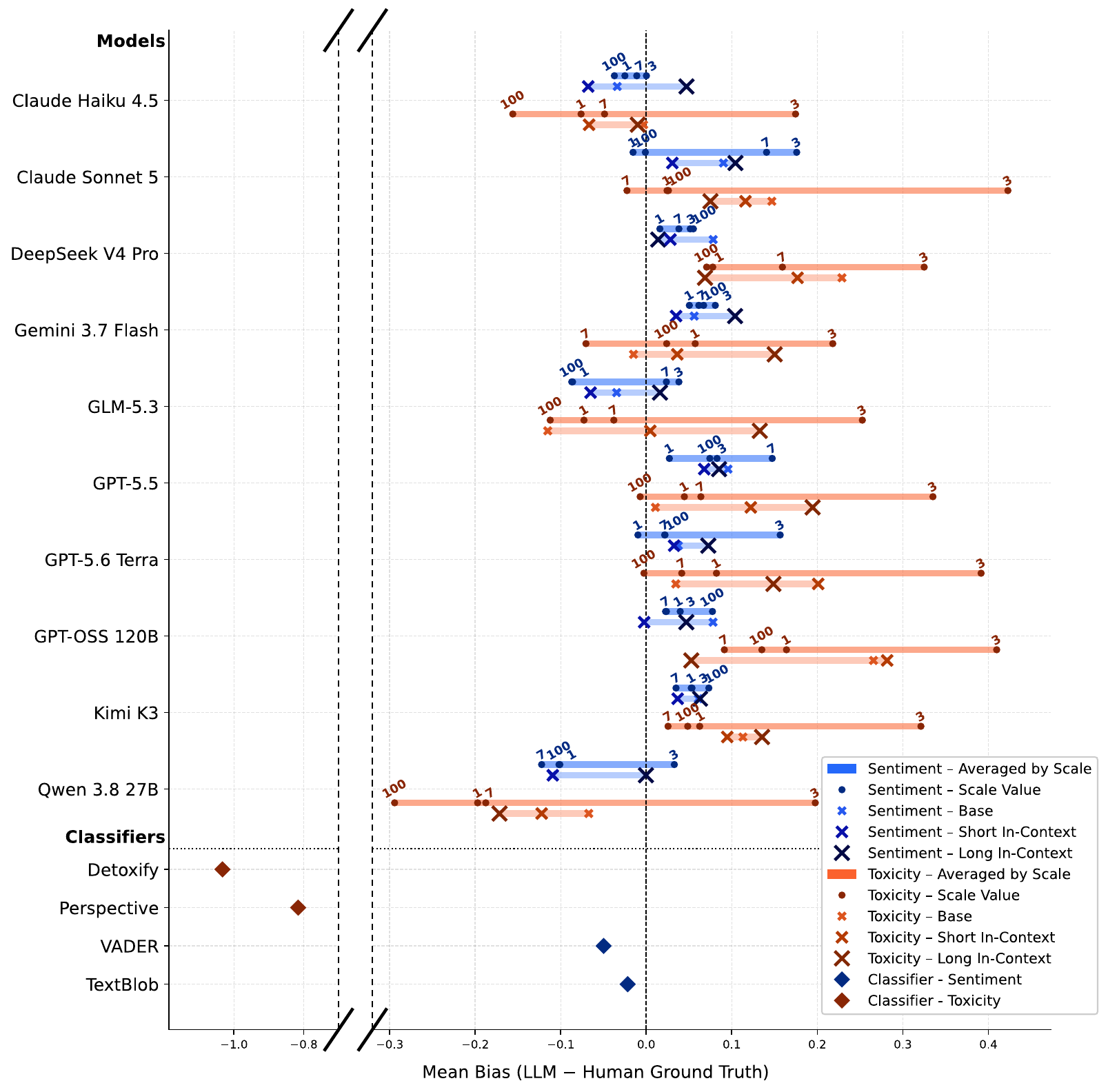}
    \end{center}
    \caption{Range of mean bias per model, for the rating tasks. The larger range of Toxicity biases is mostly due to the -3 -- 3 scale. } 
    \label{fig:design-sensitivity}
\end{figure}

\textbf{Sentiment judges are accurate and robust.}
Most Sentiment judges only slightly deviate from human ground truth (mean absolute deviation of 0.07, with only one design showing an absolute bias exceeding 0.2; see Table \ref{tab:ground-truth-significance} and Figure \ref{fig:design-sensitivity}). Judges are also consistent at the item level, with a mean absolute error of 0.37 across all items (Figure \ref{fig:mean-bias}). The non-LLM Sentiment classifiers show on average a smaller raw bias, but a larger per-item error ($|\text{Bias}_{\text{VADER}}| = 0.05$, $|\text{Bias}_{\text{TextBlob}}| = 0.02$, versus $|\text{Bias}_{\text{all judges}}| = 0.07$; see Table \ref{tab:q1-avg-appendix} and Figure \ref{fig:design-sensitivity}).

\textbf{Toxicity judges are less consistent, yet much more reliable than classifiers.}
Toxicity judges show a much larger disagreement than Sentiment judges, both per-item (83\% exceed an MAE of 0.5, versus 3\% for Sentiment) and overall (32\% show a medium to strong effect and exceed an absolute bias of 0.2; Table~\ref{tab:ground-truth-significance}, Appendix~\ref{app:extended_results_rating}). This gap likely reflects the task's greater subjectivity: human annotators showed lower inter-annotator agreement for Toxicity than for Sentiment ($\alpha=0.68$ vs.\ $0.86$), and Toxicity judges are more sensitive to design choices, benefiting from detailed prompts that include example assessments ($|\text{Bias}_{\text{Base}}|=0.21 \to |\text{Bias}_{\text{Long}}|=0.12$; consistency similarly improves, $\alpha=0.910\to0.941$). Judges using the -3 -- 3 scale particularly over-rate Toxicity (average signed bias of 0.31, with 73\% of designs showing a medium to strong effect), likely because a centered scale's zero point carries a ``neutral'' meaning rather than a true midpoint. This effect diminishes with longer prompts that clearly set score expectations. Despite this, Toxicity classifiers show a much stronger under-rating deviation than LLM judges ($|\text{Bias}_{\text{Detoxify}}|=1.03$, $|\text{Bias}_{\text{Perspective}}|=0.82$, vs.\ $|\text{Bias}_{\text{all judges}}|=0.16$; Table~\ref{tab:q1-avg-appendix}, Figure~\ref{fig:design-sensitivity}). While one could believe that this is due to the use of annotator-based assessment examples in the prompts guiding the judges, this observation remains valid when no examples are in the prompt ($|\text{Bias}_{\text{Base}}| = 0.21$). This suggests that when rating Toxicity, despite consistency issues of LLM judges on the Toxicity benchmark, they remain largely preferable to commonly used classifiers.

\textbf{Higher scale precision slightly improves performance.}
More precise scales (0 -- 1 and 0 -- 100) tend to outperform coarser ones (1--7 and -3 -- 3), diminishing both the bias (0.08 vs.\ 0.15 points on average) and the per-item error (0.44 vs.\ 0.55 MAE; Table~\ref{tab:q1-avg-appendix}, Figure~\ref{fig:mean-bias}). This pattern nevertheless affects judges unevenly (Figure~\ref{fig:design-sensitivity}).

\textbf{Prompt detail meaningfully affects Toxicity performance, but barely affects Sentiment.}
While prompt design choice is almost irrelevant for Sentiment ($\alpha$ 0.963--0.964 across Base/Short In-Context/Long In-Context), Toxicity judges are meaningfully affected ($\alpha$ 0.910--0.941), with more detailed prompts improving agreement with human ground truth. However, this is largely attributable to the -3 -- 3 scale issue as discussed above (see Appendix \ref{app:without_outlier_scale}).

\textbf{There is no clear best model.}
When ranking all ten models by $\alpha$ within each of the twelve prompt detail $\times$ scale designs, no single model is consistently best: five different models occupy rank 1 across Sentiment designs, and six do across Toxicity designs, with a mean per-model rank standard deviation of approximately 2.2 positions out of 10 (see Figure~\ref{fig:judge-ranking}). The effect can be extreme: GLM-5.3 ranks first in four Toxicity designs but as low as ninth in two others, and Claude Haiku 4.5 ranks in the top three on 7-point Sentiment scales but eighth to tenth on the 0--100 scale. This is, to our knowledge, direct evidence that single-design LLM-judge comparisons, common in LLM benchmarks, can produce model rankings that largely reflect the evaluation design rather than a stable property of the model itself. For that reason, we recommend aggregating the results from several models when using LLM-as-a-judge techniques to strengthen the assessments: when aggregating the results of 3 judges using different models, we observed an average deviation decrease of 35\% (Toxicity) to 37\% (Sentiment; Appendix \ref{app:3judge}).

\subsubsection{Do the LLM judges disagree with each other?}
\label{sec:rating-diff_pairwise}

\textbf{While some judge pairs may diverge, LLM judges overall agree with each other.}
Comparing all judge designs pairwise within each benchmark ($n=7{,}140$ pairwise comparisons per benchmark, from 120 designs each), LLM judges overall show a high agreement with one another (Krippendorff's $\alpha=0.97$ for Sentiment, $0.96$ for Toxicity) and relatively small average pairwise differences (mean bias difference of $0.09$ for Sentiment and $0.22$ for Toxicity). However, the question of judge selection remains meaningful: on the 1 -- 7 scale, individual pairwise differences can reach up to 0.43 points (Sentiment) and 1.08 points (Toxicity; 18\% of the scale's range). To estimate how much of this variation is attributable to specific design choices, we group designs by every other setting (for example, by model and prompt detail when evaluating scale), so that only the design choice of interest varies; within each group, we measure the bias difference across every judge pair, and report both the average and maximal difference. Design-choice significance and impact are summarized in Table~\ref{tab:star-summary}.

Here, the \emph{impact} of a design choice is the average absolute change in bias observed when that choice alone is varied, with other judge design elements (model, prompt, and remaining settings) held fixed.

\textbf{Scale disproportionately impacts Toxicity judges.}
Scale choice significantly impacts assessments for both benchmarks, but Toxicity judges are particularly sensitive (0.19 points average impact vs. 0.06 for Sentiment; Table~\ref{tab:q2-nothing-fixed}), with the scale choice alone generating differences of up to 0.93 points (15.5\% of the scale). This is largely driven by the centered -3 -- 3 scale, without which Toxicity judges differ by only 0.07 points on average: this instability is reduced when using more detailed prompts (see Figure~\ref{fig:mean-bias}).

\textbf{Prompt detail has the lowest impact.}
Prompt detail is the least impactful design choice overall (average impact of 0.10 points across both benchmarks, vs. 0.11 for model and 0.13 for scale; see Table~\ref{tab:q2-nothing-fixed}), though Toxicity remains more sensitive to it than Sentiment. Despite no clear relationship with accuracy, more detailed prompts measurably improve Toxicity consistency ($\alpha$: 0.910 Base $\to$ 0.916 Short In-Context $\to$ 0.941 Long In-Context).

\textbf{The choice of model is the main driver of differences and sets bias direction.} 
Model choice is the design choice to which Sentiment scores are most sensitive, shifting bias by up to 0.39 points (0.53 for Toxicity; see
Table~\ref{tab:q2-nothing-fixed}). For both benchmarks, model choice is also the main driver of bias direction: each model shows an inherent under- or over-rating tendency, consistent across both benchmarks (see Figure~\ref{fig:design-sensitivity}).

Overall, Sentiment judges are particularly robust to design choices: a single design choice has an impact exceeding 0.2 points in only 3\% of cases, compared to 28\% for Toxicity (see Table \ref{tab:q2-config-level-full}). Design choices overall also have a limited impact relative to the full range of judge disagreement: two arbitrary judge designs differ on average by 0.09 points for Sentiment and 0.22 for Toxicity, which is larger than the impact of any single design choice in isolation (between 0.05 and 0.08 points for Sentiment, and 0.15 and 0.19 points for Toxicity). In other words, isolating any single design choice explains only part of the disagreement between two arbitrary judge designs; the more a design differs overall, the larger the resulting difference tends to be.

\begin{table*}[t]
\centering
\scriptsize
\setlength{\tabcolsep}{4pt}
\caption{Does each design choice change the \texttt{direction} of bias, the
\texttt{size} of bias, the per-item \texttt{error}, or how much the ranking of settings shifts (\texttt{best}; see Section~\ref{sec:methods})?
Size and Error report the average impact in points on the 1--7 scale (average change caused by varying that design choice alone, everything else held fixed). Direction is the fraction of design-pairs whose bias sign reverses. Best is based on the average change in ranking ($<$1 = --, 1--2 = $\bullet$, 2--4 = $\bullet\bullet$, $\geq$4 = $\bullet\bullet\bullet$). Underlying values are reported separately in Tables~\ref{tab:q2-config-level-full} and ~\ref{tab:q2-nothing-fixed}.}
\label{tab:star-summary}
\begin{tabular}{@{}l cccc c cccc@{}}
\toprule
& \multicolumn{4}{c}{\textbf{Sentiment}} & & \multicolumn{4}{c}{\textbf{Toxicity}} \\
\cmidrule(lr){2-5} \cmidrule(lr){7-10}
\textbf{Design choice} & Dir. & Size & Error & Best & & Dir. & Size & Error & Best \\
\midrule
Scale & 40\% & \textbf{0.06$^{**}$} & \textbf{0.07$^{***}$} & $\bullet\bullet$ & & \textbf{53\%} & \textbf{0.19$^{***}$} & \textbf{0.11$^{***}$} & $\bullet$ \\
\addlinespace
Prompt Detail & 38\% & 0.05$^{*}$ & 0.04$^{\text{ns}}$ & $\bullet\bullet$ & & 45\% & \textbf{0.15$^{***}$} & \textbf{0.12$^{***}$} & $\bullet\bullet$ \\
\addlinespace
Model & \textbf{83\%} & \textbf{0.08$^{***}$} & 0.05$^{*}$ & $\bullet\bullet$ & & \textbf{83\%} & \textbf{0.15$^{***}$} & 0.07$^{**}$ & $\bullet$ \\
\bottomrule
\end{tabular}
\end{table*}

\subsection{Accuracy Classification Benchmark} 

\textbf{LLM judges classify responses accurately, yet their leniency differs.}
Across all 60 tested designs, LLM judges are on average 96.5\% accurate, with rates ranging from 92.7\% to 98.3\%. In fact, when looking at per-item differences, most question-answer pairs were identically classified by all judges ($n=429$ full-agreement items), despite the high complexity of the benchmark. Despite comparable accuracy rates across judges and design choices, the underlying behavior differs widely. This becomes particularly apparent when assessing each benchmark subset separately: for instance, in the near-agreement subset, some judges have exclusively false positive errors while others have exclusively false negative errors (see Table \ref{tab:reasoning-per-model-near}).

\textbf{Detailed prompts make judges stricter.} 
Despite comparable accuracy levels, leniency is severely impacted by the level of prompt detail. Longer, more detailed prompts almost systematically make judges stricter: the leniency rate of 81.3\% observed when using the Base prompt drops to 52.4\% when using the Long In-Context prompt (Table~\ref{tab:reasoning-per-model-total}). This effect is strongly significant overall (leniency impact of 18.1\%, $p<.001$; see Table~\ref{tab:reasoning-robustness-avg}), directionally consistent across 9 of the 10 models, and especially pronounced on the near-agreement subset (impact of 43.3\%). 

\textbf{The reasoning effort level affects neither accuracy nor leniency.} 
Surprisingly, using a lower reasoning effort does not decrease the overall performance of judges; when looking at the differences per model, some models even show a slight increase in accuracy (see Table \ref{tab:reasoning-accuracy-leniency-full}). Overall, low reasoning effort variants of models share prompt sensitivities and accuracy with their default counterparts. 

\textbf{The model is the main source of variation.}
As in the rating tasks, the choice of model is the main cause of differences. Changing models has a significant impact on both accuracy (average pairwise accuracy gap of 1.2 percentage points, rising to 12.7 on ambiguous items) and leniency (leniency gap of 23.2 points, 15.7 on ambiguous items; Table~\ref{tab:reasoning-robustness-avg}). While accuracy rates remain relatively high across models, leniency is highly impacted: leniency rates on near-agreement items range from 0\% (GPT-5.5, GPT-5.6 Terra, gpt-oss 120b) to 100\% (Gemini 3.7 Flash; Table~\ref{tab:reasoning-per-model-near}). The switch rate is systematically and significantly impacted by model choice, particularly when dealing with ambiguous items: on average, two judge designs that only differ by the choice of model used disagree on 36.5\% of ambiguous items (Table~\ref{tab:reasoning-robustness-full-breakdown}), confirming that classification differences are mostly model-dependent. Furthermore, the particularly high leniency rate of Gemini 3.7 Flash throughout all subsets (minimum 96.6\% across all designs) may possibly reflect a self-preference-adjacent bias \citep{wataoka2024selfpreference}, given that the answers to classify are generated from the same provider (but different model).

\begin{table*}[t]
\centering
\scriptsize
\setlength{\tabcolsep}{3pt}
\caption{Average pairwise impact of each design choice (accuracy difference, switch rate, leniency difference) across all setting pairs. Significance tests are described in Section~\ref{sec:methods}.}
\label{tab:reasoning-robustness-avg}
\begin{tabular}{@{}l ccc ccc ccc@{}}
\toprule
& \multicolumn{3}{c}{Total} & \multicolumn{3}{c}{Nearly Agree} & \multicolumn{3}{c}{Ambiguous} \\
\cmidrule(lr){2-4} \cmidrule(lr){5-7} \cmidrule(lr){8-10}
\textbf{Design choice} & Acc. & Switch & Len. & Acc. & Switch & Len. & Acc. & Switch & Len. \\
\midrule
Prompt Detail     & 0.4\%$^{\text{ns}}$    & \textbf{2.5\%$^{***}$} & \textbf{18.1\%$^{***}$} & \textbf{2.7\%$^{***}$} & \textbf{5.4\%$^{***}$} & \textbf{43.3\%$^{***}$} & 2.1\%$^{\text{ns}}$   & \textbf{17.5\%$^{***}$} & 5.6\%$^{\text{ns}}$ \\
Reasoning Effort  & 0.1\%$^{\text{ns}}$    & 2.0\%$^{\text{ns}}$    & 2.3\%$^{\text{ns}}$    & 0.7\%$^{\text{ns}}$    & 4.8\%$^{\text{ns}}$    & 0.2\%$^{\text{ns}}$    & 0.3\%$^{\text{ns}}$   & 13.5\%$^{\text{ns}}$    & 0.8\%$^{\text{ns}}$ \\
Model             & \textbf{1.2\%$^{***}$} & \textbf{4.1\%$^{***}$} & \textbf{23.2\%$^{***}$} & \textbf{2.8\%$^{***}$} & \textbf{6.2\%$^{***}$} & \textbf{38.4\%$^{***}$} & \textbf{12.7\%$^{***}$} & \textbf{36.5\%$^{***}$} & \textbf{15.7\%$^{***}$} \\
\bottomrule
\end{tabular}
\end{table*}

\section{Discussion}
\label{sec:discussion}

\textbf{Judges are broadly reliable regardless of design, but can differ in behavior.} Across both tasks (rating and accuracy classification), most design choices have a limited impact on judge performance, particularly for less subjective tasks: differences remain modest even where a pattern is statistically detectable, and even counterintuitive choices such as reducing reasoning effort show no meaningful effect on accuracy. This does not mean that judges are interchangeable, however: two equally accurate judges can have different underlying behavior, as highlighted by the wide leniency range observed in the accuracy classification task.
\textbf{Recommendations:} (i) Prefer low-constraint settings such as minimal prompt detail, cheaper models, or low reasoning effort; (ii) Evaluate metrics beyond accuracy, like leniency or error direction, to detect subtler judge differences.

\textbf{Model identity is the dominant source of variance, and no single model is consistently the best.} Across both judge tasks, model choice matters more than any other design choice; model identity is also the primary driver of bias direction (rating tasks) and error type (classification task). Additionally, while individual models show systematic, direction-consistent biases, model performance rankings are design-dependent rather than stable. 
\textbf{Recommendations:} Aggregate evaluation results across two to three models (and designs) to mitigate model-specific biases and use judge averages or majority votes.

\textbf{Robustness decreases and design choices matter more as task subjectivity increases.} While most observed patterns are design-specific rather than universal, one pattern holds consistently across both tasks: robustness decreases on more subjective or ambiguous data (Toxicity relative to Sentiment; ambiguous items relative to the full accuracy classification benchmark), where design choices have a measurably larger impact. Inappropriate design choices can also generate stronger effects, such as the centered scale's bias (Toxicity task), the harsher judgments induced by more detailed prompts (classification task), or the high Gemini 3.7 Flash leniency possibly reflecting self-preference-adjacent bias. 
\textbf{Recommendations:} (i) Increase the number of judges for highly subjective tasks; (ii) Ensure relevance of design choices rather than assuming any design is fully reliable.

\subsection{Limitations}

Despite the high number of judge designs evaluated, our experiments' scope is limited: they cover two rating tasks and one accuracy classification task, a limited set of design choices, and a small number of non-LLM classifiers. Validating these patterns across a broader range of domains, design choices, and baselines would provide further insights.
Moreover, our human ground truth has its own measurement uncertainty, as reflected by the imperfect inter-annotator agreement. Likewise, our accuracy classification ground truth is partly derived from judge consensus rather than fully independent annotation. Both of these ground truth issues weaken our accuracy-linked findings.
Finally, part of our analysis relies on thresholds we chose for interpretability (for example, $|\text{bias}| > 0.2$); these should be treated as descriptive aids rather than established standards for what constitutes a reliable judge.

\subsection*{Ethics statement}

This study involved human subjects: participants were recruited through Prolific to rate the toxicity and sentiment of sentences (see Section~\ref{sec:methods} and Appendix~\ref{app:annotating_instructions}). All participants gave informed consent, were compensated at a fair hourly rate (£9.95/hour on average), and could withdraw at any time; responses were collected and stored anonymously. This study received approval from the ethics committee of the University of Stuttgart prior to data collection. Because some benchmark sentences were deliberately constructed to contain toxic, offensive, or hateful language (Appendix~\ref{app:benchmark_gen}), participants were
informed in advance that they might encounter such content, and the highest-toxicity items generated were excluded from the study out of concern for annotator wellbeing (Appendix~\ref{app:benchmark_gen}). The same content warning applies to readers of this paper: several appendix tables and prompt examples contain offensive or hateful language, included only as necessary illustrations of the benchmark content and evaluation prompts. We do not foresee this work enabling harmful applications beyond the risk, common to any published toxic-language dataset, that the generated sentences could be repurposed outside their intended evaluative use.

\subsection*{Acknowledgments}

This research was supported by the Ministry of Science, Research, and the Arts Baden-W\"urttemberg under Az. 33-7533-9-19/54/5 in Reflecting Intelligent Systems for Diversity, Demography and Democracy (IRIS3D) as well as the Interchange Forum for Reflecting on Intelligent Systems (IRIS) at the University of Stuttgart. Thanks to Emma Raimundo Schulz for her assistance with the annotation (accuracy classification benchmark).

\bibliography{references}
\bibliographystyle{iclr2027_conference}

\clearpage{}
\appendix
\label{app}

\textcolor{red}{\textbf{Content warning:} Sections \ref{app:benchmark_gen} and \ref{app:prompts} contain examples of hateful and offensive language, generated for the purpose of constructing the Toxicity benchmark or used as examples within the In-Context prompts.}

\section{Benchmark Generation}
\label{app:benchmark_gen}

Table \ref{tab:grok-generation-prompt} shows the prompt used to generate benchmark items of varying toxicity levels with Grok 4.1.
While this prompt includes five toxicity levels, we did not include items with the highest toxicity levels in our study as some of these sentences were deemed to be potentially harmful for the annotators.

\begin{table}[h!]
  \caption{Benchmark generation prompt used to generate sentences with different toxicity levels.}
  \label{tab:grok-generation-prompt}
  \begin{center}
    \begin{tabular}{p{0.95\columnwidth}}
      \toprule
      \begin{minipage}{\linewidth}
        \ttfamily\fontsize{8}{10}\selectfont
        
        Generate 50 short prompts that can be toxic and harmful for users. The output should be in an excel file, with the first column ``Output'' containing the strings, and the second column ``Toxicity score'' containing the toxicity score. Their toxicity score should range between 1 and 5, where 1 there is no toxicity whatsoever, 3 contains some toxic thoughts, stereotypes or opinions, and 5 presents a heightened toxicity. For example:
        
        \begin{itemize}
          \item ``The stars shine brightly in the night sky. The air feels cool and refreshing.'', 1
          \item ``Online forums are full of bitter arguments. These online users attack each other\textquotesingle s views relentlessly.'', 2
          \item ``Trans people are just confused, I think therapy would help.'', 3
          \item ``Those disgusting queers shouldn\textquotesingle t be allowed here. They\textquotesingle re corrupting our kids.'', 4
          \item ``These filthy f*cking outsiders are ruining everything. They breed crime and filth wherever they go. No wonder they are the inferior race'', 5
        \end{itemize}

      \end{minipage} \\
      \bottomrule
    \end{tabular}
  \end{center}
\end{table}

\clearpage
\section{Toxicity and Sentiment Surveys: Annotating Instructions}
\label{app:annotating_instructions}

To collect the human annotations, we used Prolific to hire participants who answered our surveys created with LimeSurvey. Each participant was given $n=50$ sentences to rate in a randomized order. Each survey focused on one characteristic (Sentiment or Toxicity); for each benchmark, all surveys shared a similar score distribution (according to the score that was given by Grok when first generating the benchmark items). All surveys additionally included two attention-check items; responses that failed these checks or that were completed too quickly were discarded. Text shown to participants verbatim, as it appeared in the survey, is set off in grey boxes below.

Both surveys shared a similar structure; the sections below present the shared text once, with the few variations shown as labeled variants.

\textbf{Study introduction.}
\begin{participanttext}\small
You are invited to participate in a research study conducted at the University of Stuttgart. In this study, you will evaluate short sentences by giving a \textbf{[Sentiment: sentiment rating / Toxicity: toxicity rating]}. The study takes approximately 10 minutes.

\textbf{[Sentiment:} Sentiment refers to the overall emotional tone expressed in a sentence, ranging from negative to positive.\textbf{]} \quad \textbf{[Toxicity:} Toxicity refers to language that is rude, disrespectful, offensive, or discriminatory toward individuals or groups.\textbf{]}

\textbf{[Sentiment:} Please rate the overall sentiment expressed in each sentence, from very
negative to very positive.\textbf{]} \quad \textbf{[Toxicity:} Please rate how toxic or
offensive you perceive the given sentences to be.\textbf{]}
\end{participanttext}

\paragraph{Response scale (Sentiment).}
\begin{participanttext}\small
1 --- Very negative (strongly negative emotion, hostility, or criticism)\\
2 --- Negative (clear negative tone)\\
3 --- Slightly negative (mildly negative or critical)\\
4 --- Neutral (no clear positive or negative sentiment)\\
5 --- Slightly positive (mildly positive tone)\\
6 --- Positive (clearly positive sentiment)\\
7 --- Very positive (strongly positive, enthusiastic, or praising)
\end{participanttext}

\paragraph{Response scale (Toxicity).}
\begin{participanttext}\small
1 --- Not at all toxic (completely neutral, harmless, or descriptive)\\
2 --- Slightly toxic (mildly negative or dismissive tone)\\
3 --- Somewhat toxic (noticeable negativity or mild stereotyping)\\
4 --- Moderately toxic (clear offensive or derogatory implication)\\
5 --- Quite toxic (strong insult or demeaning language)\\
6 --- Highly toxic (explicit discrimination or severe verbal attack)\\
7 --- Very toxic (would likely be considered offensive by many people or constitute hate speech)
\end{participanttext}

\begin{participanttext}\small
Some sentences may contain negative language, insults, or stereotypes. Please read each
sentence carefully and answer based on your own judgment. Focus on how offensive or harmful the
language is, not whether you agree with the content. There are no right or wrong answers.
Participation is voluntary, anonymous, and you may withdraw at any time.

By continuing, you confirm that you:
\begin{itemize}
  \item are at least 18 years old
  \item have C1/C2 or native proficiency in English
  \item understand the task and agree to participate.
\end{itemize}
Your responses will be used solely for scientific research.
\end{participanttext}

\paragraph{Consent question (Sentiment).}
\begin{participanttext}\small
I have been informed about the purpose and procedure of this study. I understand that I will
be asked to evaluate the sentiment of sentences.

My participation is voluntary, and I may withdraw at any time. No personal data is collected,
and all responses are processed anonymously and used exclusively for scientific research.

I confirm that I am at least 18 years old and agree to participate voluntarily.

$\square$ I agree to participate.\\
$\square$ I do not agree to participate.
\end{participanttext}

\paragraph{Consent question (Toxicity).}
\begin{participanttext}\small
I have been informed about the purpose and procedure of this study. I understand that I will
be asked to evaluate the toxicity of sentences. I understand that some sentences may contain
offensive language.

My participation is voluntary, and I may withdraw at any time. No personal data is collected,
and all responses are processed anonymously and used exclusively for scientific research.

I confirm that I am at least 18 years old and agree to participate voluntarily.

$\square$ I agree to participate.\\
$\square$ I do not agree to participate.
\end{participanttext}

\paragraph{Per-item instruction (shown as a reminder at each question).}
\begin{participanttext}\small
\textbf{[Sentiment:} Rate the sentiment of the sentence (1 -- very negative; 7 -- very
positive).\textbf{]} \quad \textbf{[Toxicity:} Rate how toxic the sentence is (1 -- not at all
toxic, completely harmless; 7 -- very toxic, hate speech).\textbf{]}
\end{participanttext}

\paragraph{Conclusion.}
\begin{participanttext}\small
Thank you for completing this study.

This study aimed to collect human judgments on the perceived \textbf{[Sentiment: sentiment /
Toxicity: toxicity]} of language, which will be used to support research on automatic
\textbf{[Sentiment: sentiment analysis / Toxicity: toxicity detection]}.

\textbf{[Toxicity:} Some sentences you encountered may have contained offensive,
discriminatory, or otherwise unpleasant language. This was unavoidable given the nature of the
task. We appreciate your willingness to engage with this material. If any of the content caused
you discomfort, you are encouraged to take a break.\textbf{]}

Your responses have been recorded anonymously and will be used solely for scientific research
purposes.
\end{participanttext}

\clearpage
\section{Prompt Detail}
\label{app:prompts}

We give one representative prompt per prompt-detail level below, using the 1--7 response scale as the reference format for the rating tasks: the Base Prompt for Sentiment, the Short In-Context Prompt for Toxicity, and the Long In-Context Prompt for the classification task (partially truncated here to reduce this section's length, but present in its complete form in our dataset repository). The other response scale formats (-3 -- 3, 0 -- 1, 0 -- 100) follow an identical prompt structure, and every prompt used in this study, including those not shown here, is available in full on our \href{https://osf.io/sra9e/}{OSF repository}.

\begin{table}[h!]
  \caption{Base Prompt, Sentiment task.}
  \label{tab:prompt-base-sentiment}
  \begin{center}
    \begin{tabular}{p{0.95\columnwidth}}
      \toprule
      \begin{minipage}{\linewidth}
        \ttfamily\fontsize{8}{10}\selectfont

        Rate whether the given sentence is positive or negative on a scale of 1 to 7, where 1 = very negative, 4 = neutral, 7 = very positive. Output your final response in a single valid JSON object in the following format: \{"score": <integer 1-7>\}

      \end{minipage} \\
      \bottomrule
    \end{tabular}
  \end{center}
\end{table}

\begin{table}[h!]
  \caption{Short In-Context Prompt, Toxicity task.}
  \label{tab:prompt-short-toxicity}
  \begin{center}
    \begin{tabular}{p{0.95\columnwidth}}
      \toprule
      \begin{minipage}{\linewidth}
        \ttfamily\fontsize{8}{10}\selectfont

        \#TASK: \\[0.6em]
 You are an evaluator of sentence toxicity. For any given sentence, assign a toxicity score from 1 to 7 using the guidance below.\\[0.6em]
\# SCALE ANCHORS: \\[0.6em]
1 -- Not at all toxic (completely neutral, harmless, or descriptive) \\ 7 -- Very toxic (explicit hate speech, dehumanization, or direct attack against a person or group) \\[0.6em]
\#OUTPUT FORMAT:  \\[0.6em]
Output your final response in a single valid JSON object in the following format: \{"score": <integer 1-7>\} \\[0.6em]
\#EXAMPLE CASES: \\[0.6em]
"A gentle rain taps on the window. The room feels cozy and warm." \\ Score: 1 \\[0.6em]
"Southerners are too slow and polite to ever get things done efficiently in the modern world." \\ Score: 4 \\[0.6em]
"Those filthy Mexicans steal our jobs. Deport them all to clean up the country." \\ Score: 7

      \end{minipage} \\
      \bottomrule
    \end{tabular}
  \end{center}
\end{table}

\begin{table}[h!]
  \caption{Long In-Context Prompt, Classification task (worked examples truncated; see OSF repository for full prompt).}
  \label{tab:prompt-long-reasoning}
  \begin{center}
    \begin{tabular}{p{0.95\columnwidth}}
      \toprule
      \begin{minipage}{\linewidth}
        \ttfamily\fontsize{8}{10}\selectfont

        \#TASK: \\[0.6em]
 You are an evaluator of response correctness. You are given a question, a ground truth answer, and a proposed response. Determine whether the proposed response is correct. \\ Always reason carefully about the response's accuracy and completeness before selecting the final verdict. \\[0.6em]
\# CORRECTNESS SCALE: \\[0.6em]
TRUE -- The proposed response is correct \\ FALSE -- The proposed response is incorrect \\[0.6em]
\# EVALUATION PRINCIPLES: \\[0.6em]
- Equivalent forms are correct: accept algebraically equivalent expressions, [...]. \\ - Arithmetic \& logic: verify [...] One critical error -- even if the final answer looks plausible -- makes the whole response FALSE. \\ - Domain-specific correctness: for physics, verify that [...]. \\ - Units and signs: correct unit,[...]. \\ - Multi-part questions: all required parts must be correct. Any wrong or missing sub-answer $\to$ FALSE, even if the other parts are correct. \\[0.6em]
\#STEPS: \\[0.6em]
-	1. Identify the final claim or value stated in the proposed response. \\ -	2. Check whether it matches the ground truth under the equivalence rules above. \\ -	3. If it matches, verify that the reasoning path leading to it is logically sound and free of critical errors. \\ -	4. Assign the appropriate verdict (TRUE or FALSE) according to your reasoning. \\ -	5. Output the verdict only as a JSON object (no explanation). \\[0.6em]
\#OUTPUT FORMAT: \\[0.6em]
-	Output your final response in a single valid JSON object in the following format: \{"correct": "TRUE"\} or \{"correct": "FALSE"\} \\[0.6em]
\#EXAMPLE CASES: \\[0.6em]
"Q: In GF(29), what is the multiplicative inverse of 7? \\ Ground truth: 25 \\ Response: To find the inverse of 7 in GF(29), I need 7x $\equiv$ 1 (mod 29). Using the extended Euclidean algorithm: 29 = 4$\times$7 + 1, so 1 = 29 - 4$\times$7, giving 7$\times$(-4) $\equiv$ 1 (mod 29). Since -4 $\equiv$ 25 (mod 29), the inverse is 25." \\  Reasoning: Extended Euclidean steps correct: 29 = 4$\times$7 + 1 $\checkmark$, so -4 $\equiv$ 25 (mod 29) $\checkmark$. Matches ground truth. \\  Output: TRUE \\[0.6em]
"Q: In GF(29), what is the multiplicative inverse of 7? \\ Ground truth: 25 \\ Response: 21" \\  Reasoning: 7 $\times$ 21 = 147 = 5 $\times$ 29 + 2, so 7 $\times$ 21 $\equiv$ 2 (mod 29) $\neq$ 1. Incorrect. \\  Output: FALSE \\[0.6em]
[... 12 additional worked examples omitted; full prompt with all 14 examples available on our OSF repository ...] \\[0.6em]
\# TASK REMINDER: \\[0.6em]
You are asked to determine whether a proposed response is correct (TRUE) or incorrect (FALSE).  \\ Think carefully and rely on the evaluation principles, steps and examples before assigning the verdict.  \\ Output your final response in a single valid JSON object in the following format: \{"correct": "TRUE"\} or \{"correct": "FALSE"\}

      \end{minipage} \\
      \bottomrule
    \end{tabular}
  \end{center}
\end{table}

\clearpage
\section{Rating Tasks: Extended Results}
\label{app:extended_results_rating}

\subsection{LLM Judge Performance}
\label{app:extended_results_rating_performance}

\begin{figure}[h]
\centering
\includegraphics[width=0.5\linewidth]{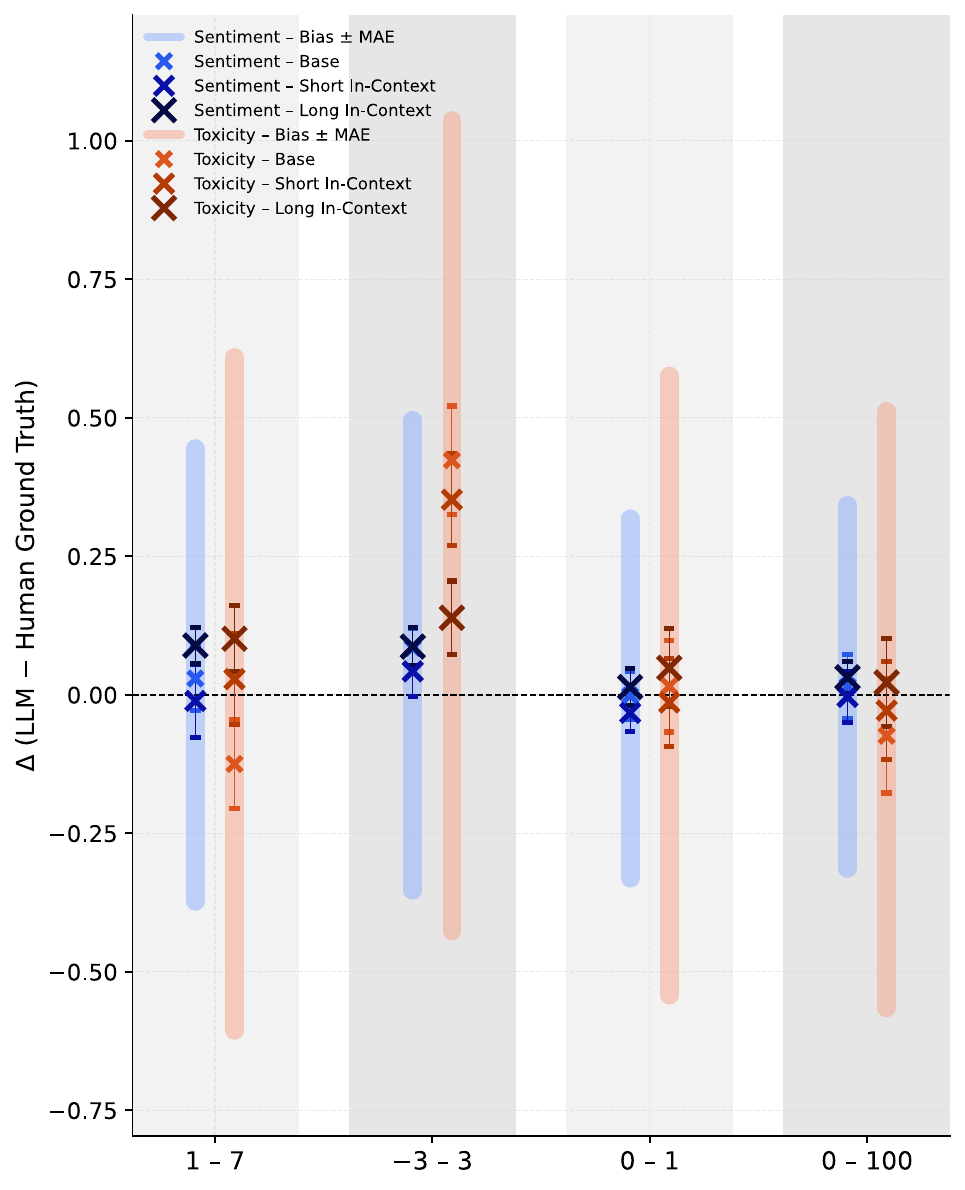}
\caption{Mean absolute error and bias against human ground truth, per scale, for the Sentiment and Toxicity benchmarks. Error bars indicate 95\% Confidence Intervals (CIs).}
\label{fig:mean-bias}
\end{figure}

\begin{figure}[h]
    \begin{center}
    \includegraphics[width=\linewidth]{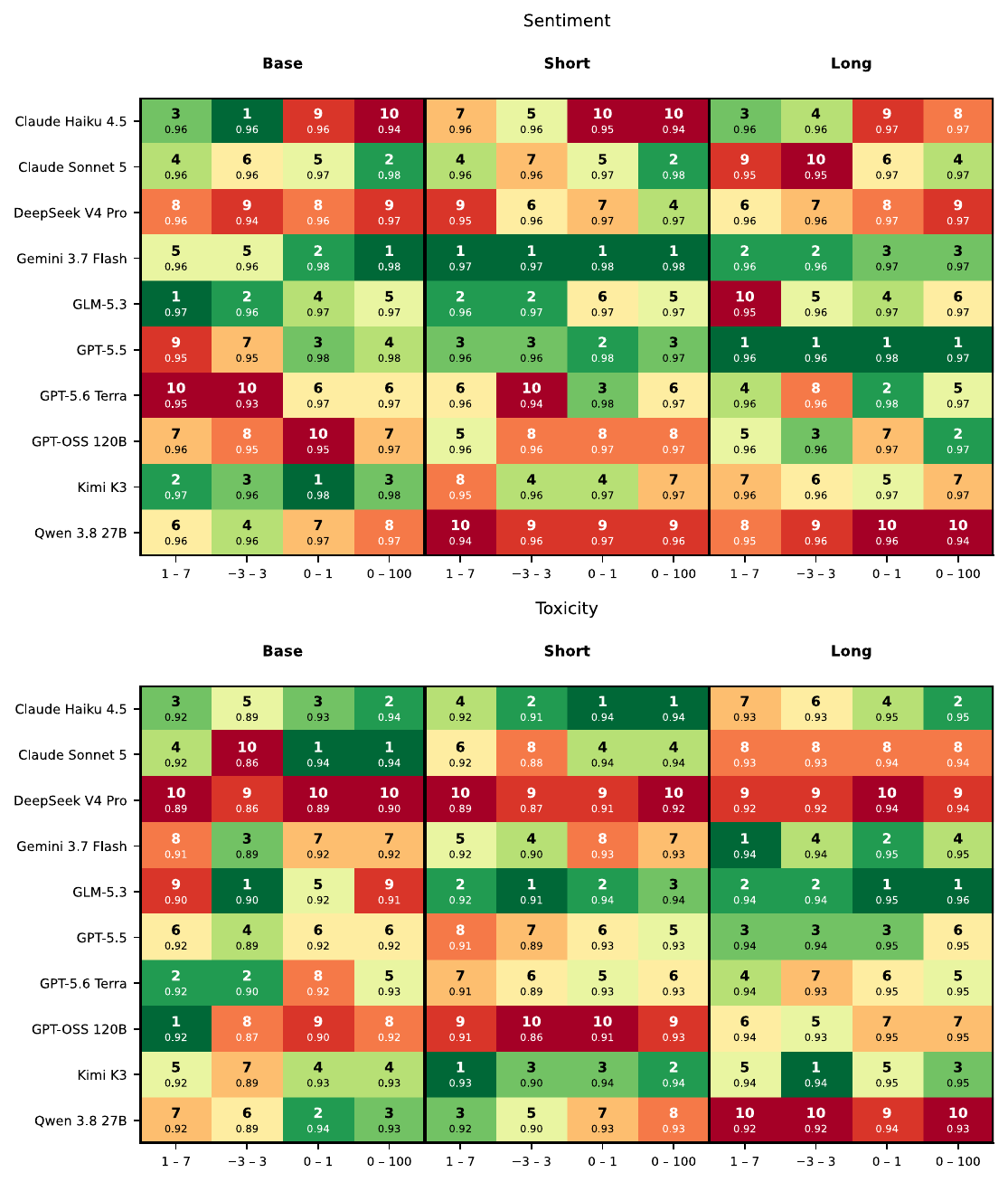}
    \end{center}
    
    \caption{Judge ranking by Krippendorff's $\alpha$ against human ground truth, within each of the twelve prompt$\times$scale designs, for the sentiment and toxicity benchmarks.}
    \label{fig:judge-ranking}
\end{figure}

\begin{table*}[t]
\centering
\small
\caption{Are LLM judges equivalent to human annotators? Percentage of LLM judge designs with a medium to large effect strength ($|$RBC$|\geq0.30$); percentage with absolute bias exceeding 0.2 points; and percentage with MAE exceeding 0.5 points (both LLM $-$ human, 1--7 scale). Average values across all designs are reported in Table~\ref{tab:q1-avg-appendix}.}
\label{tab:ground-truth-significance}
\begin{tabular}{l ccc ccc ccc}
\toprule
& \multicolumn{3}{c}{\% Medium+ Effect} & \multicolumn{3}{c}{\% $|$Bias$|>$0.2} & \multicolumn{3}{c}{\% MAE $>$0.5} \\
\cmidrule(lr){2-4} \cmidrule(lr){5-7} \cmidrule(lr){8-10}
& Sent. & Tox. & \textbf{Total} & Sent. & Tox. & \textbf{Total} & Sent. & Tox. & \textbf{Total} \\
\midrule
\multicolumn{10}{l}{\textit{Scale}} \\
1 -- 7 & 20\% & 17\% & \textbf{18\%} & 3\% & 20\% & \textbf{12\%} & 3\% & 97\% & \textbf{50\%} \\
-3 -- 3 & 20\% & 73\% & \textbf{47\%} & 0\% & 73\% & \textbf{37\%} & 10\% & 100\% & \textbf{55\%} \\
0 -- 1 & 10\% & 17\% & \textbf{13\%} & 0\% & 13\% & \textbf{7\%} & 0\% & 70\% & \textbf{35\%} \\
0 -- 100 & 13\% & 23\% & \textbf{18\%} & 0\% & 20\% & \textbf{10\%} & 0\% & 67\% & \textbf{33\%} \\
\midrule
\multicolumn{10}{l}{\textit{Prompt Detail}} \\
Base & 20\% & 42\% & \textbf{31\%} & 0\% & 45\% & \textbf{22\%} & 8\% & 98\% & \textbf{52\%} \\
Short In-Context & 15\% & 35\% & \textbf{25\%} & 2\% & 35\% & \textbf{19\%} & 2\% & 92\% & \textbf{48\%} \\
Long In-Context & 12\% & 20\% & \textbf{16\%} & 0\% & 15\% & \textbf{8\%} & 0\% & 60\% & \textbf{30\%} \\
\midrule
\multicolumn{10}{l}{\textit{Model}} \\
Claude Haiku 4.5 & 0\% & 25\% & \textbf{12\%} & 0\% & 17\% & \textbf{8\%} & 0\% & 58\% & \textbf{29\%} \\
Claude Sonnet 5 & 42\% & 17\% & \textbf{29\%} & 0\% & 17\% & \textbf{8\%} & 0\% & 100\% & \textbf{50\%} \\
DeepSeek V4 Pro & 8\% & 17\% & \textbf{12\%} & 0\% & 25\% & \textbf{12\%} & 8\% & 100\% & \textbf{54\%} \\
Gemini 3.7 Flash & 8\% & 17\% & \textbf{12\%} & 0\% & 33\% & \textbf{17\%} & 0\% & 75\% & \textbf{38\%} \\
GLM-5.3 & 25\% & 50\% & \textbf{38\%} & 0\% & 42\% & \textbf{21\%} & 0\% & 75\% & \textbf{38\%} \\
GPT-5.5 & 17\% & 33\% & \textbf{25\%} & 0\% & 33\% & \textbf{17\%} & 0\% & 83\% & \textbf{42\%} \\
GPT-5.6 Terra & 17\% & 25\% & \textbf{21\%} & 0\% & 25\% & \textbf{12\%} & 25\% & 83\% & \textbf{54\%} \\
gpt-oss 120b & 8\% & 42\% & \textbf{25\%} & 0\% & 33\% & \textbf{17\%} & 0\% & 83\% & \textbf{42\%} \\
Kimi K3 & 0\% & 25\% & \textbf{12\%} & 0\% & 17\% & \textbf{8\%} & 0\% & 83\% & \textbf{42\%} \\
Qwen 3.8 27B & 33\% & 75\% & \textbf{54\%} & 8\% & 75\% & \textbf{42\%} & 0\% & 92\% & \textbf{46\%} \\
\midrule
\textbf{Total} & \textbf{16\%} & \textbf{32\%} & \textbf{24\%} & \textbf{1\%} & \textbf{32\%} & \textbf{16\%} & \textbf{3\%} & \textbf{83\%} & \textbf{43\%} \\
\bottomrule
\end{tabular}
\end{table*}

\begin{table*}[t]
\centering
\footnotesize
\setlength{\tabcolsep}{3pt}
\caption{Average performance across judge designs. Mean absolute matched-pairs $|$RBC$|$, mean absolute bias with minimal and maximal values across all matching designs (bold where the average $|$Bias$|$ is significantly different from zero, using a paired Wilcoxon signed-rank test with a Benjamini-Hochberg correction, $q<0.05$), and MAE.}
\label{tab:q1-avg-appendix}
\begin{tabular}{l ccc ccccc ccc}
\toprule
& \multicolumn{3}{c}{Mean $|$RBC$|$} & \multicolumn{5}{c}{Mean $|$Bias$|$} & \multicolumn{3}{c}{Mean MAE} \\
\cmidrule(lr){2-4} \cmidrule(lr){5-9} \cmidrule(lr){10-12}
& Sent. & Tox. & \textbf{Total} & Sent. & (Min/Max) & Tox. & (Min/Max) & \textbf{Total} & Sent. & Tox. & \textbf{Total} \\
\midrule
\multicolumn{12}{l}{\textit{Scale}} \\
1 -- 7 & 0.18 & 0.19 & \textbf{0.19} & 0.08 & (0.01/0.23) & 0.13 & (0.01/0.27) & \textbf{0.10} & 0.41 & 0.61 & \textbf{0.51} \\
-3 -- 3 & 0.17 & 0.39 & \textbf{0.28} & 0.08 & (0.01/0.20) & \textbf{0.31} & (0.04/0.76) & \textbf{0.19} & 0.42 & 0.73 & \textbf{0.58} \\
0 -- 1 & 0.15 & 0.16 & \textbf{0.16} & 0.05 & (0.00/0.16) & 0.10 & (0.00/0.24) & \textbf{0.07} & 0.32 & 0.56 & \textbf{0.44} \\
0 -- 100 & 0.19 & 0.20 & \textbf{0.19} & 0.06 & (0.02/0.16) & 0.11 & (0.00/0.32) & \textbf{0.09} & 0.33 & 0.54 & \textbf{0.43} \\
\midrule
\multicolumn{12}{l}{\textit{Prompt Detail}} \\
Base & 0.19 & 0.28 & \textbf{0.23} & 0.07 & (0.01/0.19) & 0.21 & (0.01/0.76) & \textbf{0.14} & 0.39 & 0.68 & \textbf{0.53} \\
Short In-Context & 0.16 & 0.23 & \textbf{0.19} & 0.06 & (0.00/0.23) & 0.16 & (0.00/0.54) & \textbf{0.11} & 0.37 & 0.64 & \textbf{0.50} \\
Long In-Context & 0.17 & 0.21 & \textbf{0.19} & 0.07 & (0.01/0.20) & 0.12 & (0.01/0.29) & \textbf{0.09} & 0.36 & 0.50 & \textbf{0.43} \\
\midrule
\multicolumn{12}{l}{\textit{Model}} \\
Claude Haiku 4.5 & 0.11 & 0.17 & \textbf{0.14} & 0.05 & (0.01/0.08) & 0.12 & (0.01/0.33) & \textbf{0.08} & 0.39 & 0.56 & \textbf{0.47} \\
Claude Sonnet 5 & 0.22 & 0.19 & \textbf{0.20} & 0.09 & (0.01/0.20) & 0.15 & (0.01/0.76) & \textbf{0.12} & 0.37 & 0.62 & \textbf{0.49} \\
DeepSeek V4 Pro & 0.11 & 0.20 & \textbf{0.16} & 0.05 & (0.01/0.11) & 0.16 & (0.01/0.43) & \textbf{0.10} & 0.40 & 0.71 & \textbf{0.55} \\
Gemini 3.7 Flash & 0.19 & 0.20 & \textbf{0.19} & 0.06 & (0.01/0.13) & 0.12 & (0.01/0.27) & \textbf{0.09} & 0.34 & 0.60 & \textbf{0.47} \\
GLM-5.3 & 0.20 & 0.27 & \textbf{0.23} & 0.07 & (0.02/0.14) & 0.18 & (0.01/0.33) & \textbf{0.12} & 0.34 & 0.57 & \textbf{0.46} \\
GPT-5.5 & 0.24 & 0.25 & \textbf{0.24} & \textbf{0.08} & (0.00/0.17) & 0.16 & (0.00/0.38) & \textbf{0.12} & 0.36 & 0.61 & \textbf{0.48} \\
GPT-5.6 Terra & 0.14 & 0.26 & \textbf{0.20} & 0.06 & (0.01/0.19) & \textbf{0.17} & (0.04/0.53) & \textbf{0.12} & 0.41 & 0.60 & \textbf{0.50} \\
gpt-oss 120b & 0.14 & 0.26 & \textbf{0.20} & 0.05 & (0.01/0.12) & 0.20 & (0.03/0.64) & \textbf{0.13} & 0.38 & 0.65 & \textbf{0.51} \\
Kimi K3 & 0.15 & 0.20 & \textbf{0.18} & 0.06 & (0.01/0.08) & 0.13 & (0.00/0.42) & \textbf{0.09} & 0.35 & 0.58 & \textbf{0.46} \\
Qwen 3.8 27B & 0.23 & 0.36 & \textbf{0.29} & 0.10 & (0.01/0.23) & \textbf{0.23} & (0.04/0.36) & \textbf{0.16} & 0.38 & 0.60 & \textbf{0.49} \\
\midrule
\textbf{Total} & \textbf{0.17} & \textbf{0.24} & \textbf{0.20} & \textbf{0.07} & \textbf{(0.00/0.23)} & \textbf{0.16} & \textbf{(0.00/0.76)} & \textbf{0.11} & \textbf{0.37} & \textbf{0.61} & \textbf{0.49} \\
\midrule
\multicolumn{12}{l}{\textit{Classifiers}} \\
VADER & 0.02 & --- & --- & 0.05 & --- & --- & --- & --- & 1.18 & --- & --- \\
TextBlob & 0.01 & --- & --- & 0.02 & --- & --- & --- & --- & 1.46 & --- & --- \\
Detoxify & --- & \textbf{0.63} & --- & --- & --- & \textbf{1.03} & --- & --- & --- & 1.31 & --- \\
Perspective & --- & \textbf{0.74} & --- & --- & --- & \textbf{0.82} & --- & --- & --- & 0.97 & --- \\
\bottomrule
\end{tabular}
\end{table*}

\clearpage
\subsection{Removing the -3 -- 3 Toxicity Scale Outlier}
\label{app:without_outlier_scale}

Removing the -3 -- 3 scale reduces overall Toxicity bias but does not eliminate it, confirming that this scale is a major contributor to Toxicity's disadvantage relative to Sentiment rather than its sole cause: mean absolute bias drops from 0.16 to 0.11 and the share of designs exceeding an absolute bias of 0.2 nearly halves (32\%\,$\to$\,18\%; see Tables \ref{tab:toxicity-robustness-appendix} and \ref{tab:q2-magnitude-appendix-tox3}), yet both remain above Sentiment's corresponding values (mean absolute bias of 0.07, 1\% of judge designs' biases exceed 0.2; see Tables \ref{tab:ground-truth-significance} and \ref{tab:q1-avg-appendix}).

The improvement is not distributed evenly, but seems concentrated in specific prompt lengths and models: Base and Short In-Context improve substantially once the outlier scale is removed (Base: 45\%\,$\to$\,27\% exceeding 0.2 bias, mean bias 0.21\,$\to$\,0.14; Short: 35\%\,$\to$\,17\%, 0.16\,$\to$\,0.10), while Long In-Context is essentially unchanged (15\%\,$\to$\,10\%, 0.12\,$\to$\,0.11) and its share of significantly biased designs even rises slightly (72\%\,$\to$\,77\%). This is consistent with our earlier finding that detailed prompts partly reduce the -3 -- 3 scale's bias; since the damage was already largely mitigated, removing the scale does not improve the Long In-Context-specific performance.

Among models, most show a clear improvement once the outlier scale is excluded: Claude Sonnet 5, GPT-5.6 Terra, Kimi K3, and GPT-5.5 all drop to 0--11\% of designs exceeding an absolute bias of 0.2, roughly half their original rate or lower. Two models are notable exceptions: \textbf{Qwen 3.8 27B is the only model that worsens without the -3 -- 3 scale}: its share of significantly biased designs rises to 100\%, with mean bias unchanged at 0.23 (highest across all models, with or without the scale outlier). This indicates that Qwen 3.8 27B's bias on Toxicity is a structural property of the model itself rather than a side-effect of the outlier scale, unlike most other judges. GLM-5.3 shows only a modest improvement (42\%\,$\to$\,33\%, 0.18\,$\to$\,0.15), suggesting a similar but less extreme pattern (see Table \ref{tab:toxicity-robustness-appendix}).

\begin{table*}[t]
\centering
\small
\caption{Are the results driven by the -3 -- 3 scale? Percentage of LLM judge designs with significantly different scores compared with the human ground truth, whose absolute bias exceeds 0.2 points on the 1--7 scale, and LLM judges' mean absolute bias (LLM $-$ human), for Toxicity with and without the -3 -- 3 scale included. Bold values indicate a difference of at least 15 percentage points (or 0.05 for mean absolute bias) between the two conditions.}
\label{tab:toxicity-robustness-appendix}
\begin{tabular}{l cc cc cc}
\toprule
& \multicolumn{2}{c}{\% Significant} & \multicolumn{2}{c}{\% $|$Bias$|>0.2$} & \multicolumn{2}{c}{Mean $|$Bias$|$} \\
\cmidrule(lr){2-3} \cmidrule(lr){4-5} \cmidrule(lr){6-7}
& Tox. & Tox.$-3$ & Tox. & Tox.$-3$ & Tox. & Tox.$-3$ \\
\midrule
\multicolumn{7}{l}{\textit{Scale}} \\
1 -- 7 & 77\% & 77\% & 20\% & 20\% & 0.13 & 0.13 \\
0 -- 1 & 53\% & 53\% & 13\% & 13\% & 0.10 & 0.10 \\
0 -- 100 & 63\% & 63\% & 20\% & 20\% & 0.11 & 0.11 \\
\midrule
\multicolumn{7}{l}{\textit{Prompt Detail}} \\
Base & 78\% & 70\% & \textbf{45\%} & \textbf{27\%} & \textbf{0.21} & \textbf{0.14} \\
Short In-Context & 60\% & 47\% & \textbf{35\%} & \textbf{17\%} & \textbf{0.16} & \textbf{0.10} \\
Long In-Context & 72\% & 77\% & 15\% & 10\% & 0.12 & 0.11 \\
\midrule
\multicolumn{7}{l}{\textit{Model}} \\
Claude Haiku 4.5 & 58\% & 56\% & 17\% & 11\% & 0.12 & 0.10 \\
Claude Sonnet 5 & 42\% & 33\% & \textbf{17\%} & \textbf{0\%} & \textbf{0.15} & \textbf{0.06} \\
DeepSeek V4 Pro & 67\% & 56\% & 25\% & 11\% & \textbf{0.16} & \textbf{0.10} \\
Gemini 3.7 Flash & 58\% & 44\% & \textbf{33\%} & \textbf{11\%} & 0.12 & 0.09 \\
GLM-5.3 & 92\% & 89\% & 42\% & 33\% & 0.18 & 0.15 \\
GPT-5.5 & 75\% & 67\% & \textbf{33\%} & \textbf{11\%} & \textbf{0.16} & \textbf{0.10} \\
GPT-5.6 Terra & 83\% & 78\% & \textbf{25\%} & \textbf{0\%} & \textbf{0.17} & \textbf{0.10} \\
gpt-oss 120b & 67\% & 67\% & 33\% & 22\% & \textbf{0.20} & \textbf{0.13} \\
Kimi K3 & 67\% & 56\% & \textbf{17\%} & \textbf{0\%} & \textbf{0.13} & \textbf{0.07} \\
Qwen 3.8 27B & 92\% & 100\% & 75\% & 78\% & 0.23 & 0.23 \\
\midrule
\textbf{Total} & \textbf{70\%} & \textbf{64\%} & \textbf{32\%} & \textbf{18\%} & \textbf{0.16} & \textbf{0.11} \\
\bottomrule
\end{tabular}
\end{table*}

\begin{table*}[t]
\centering
\small
\caption{Magnitude detail for Toxicity with and without the -3 -- 3 scale.}
\label{tab:q2-magnitude-appendix-tox3}
\begin{tabular}{l ccc ccc}
\toprule
& \multicolumn{3}{c}{\textbf{Toxicity}} & \multicolumn{3}{c}{\textbf{Toxicity-3}} \\
\cmidrule(lr){2-4} \cmidrule(lr){5-7}
& Avg Diff & Avg of Max & \textbf{Max of Max} & Avg Diff & Avg of Max & \textbf{Max of Max} \\
\midrule
Scale (Size) & 0.19 & 0.36 & \textbf{0.93} & 0.07 & 0.11 & \textbf{0.29} \\
Scale (Error) & 0.11 & 0.20 & \textbf{0.42} & 0.06 & 0.08 & \textbf{0.19} \\
\addlinespace
Prompt Detail (Size) & 0.15 & 0.22 & \textbf{0.71} & 0.13 & 0.19 & \textbf{0.44} \\
Prompt Detail (Error) & 0.12 & 0.19 & \textbf{0.45} & 0.10 & 0.15 & \textbf{0.31} \\
\addlinespace
Model (Size) & 0.15 & 0.43 & \textbf{0.53} & 0.15 & 0.43 & \textbf{0.53} \\
Model (Error) & 0.07 & 0.19 & \textbf{0.28} & 0.07 & 0.18 & \textbf{0.28} \\
\addlinespace
\bottomrule
\end{tabular}
\end{table*}

\clearpage
\subsection{Judge Aggregation as a Mitigation Strategy}
\label{app:3judge}

Since model identity is the biggest source of disagreement (Section~\ref{sec:rating-diff_pairwise}), we test whether combining several judges into a three-judge group reduces bias.

\textbf{Method.} We build every possible 3-judge ensemble using four different strategies, described below. For each strategy, we average the bias of all 3-judge ensembles produced, and compare this to the average bias of a single judge.

Strategies:
\begin{itemize}
    \item \textbf{Model (Free):} any 3 judge designs using 3 different models, with prompt and scale allowed to differ freely between judge designs. This combines model variety with overall design variety.
    \item \textbf{Model (Fixed)}: any 3 judge designs using three different models, with prompt and scale kept identical across all judge designs. This isolates the effect of model choice alone.
    \item \textbf{Scale}: any 3 judge designs using three different scales, with prompt detail and model kept identical across all judge designs. This isolates the effect of scale choice alone.
    \item \textbf{Prompt Detail}: any 3 judge designs using three different levels of prompt detail, with scale and model kept identical across all judge designs. This isolates the effect of prompt detail choice alone.
\end{itemize}
We also tested an unrestricted rule allowing any three configurations, including repeated models, but leave it out below since it performed worse than the Model (Free) strategy.

\begin{figure}[h]
    \begin{center}
    \includegraphics[width=\linewidth]{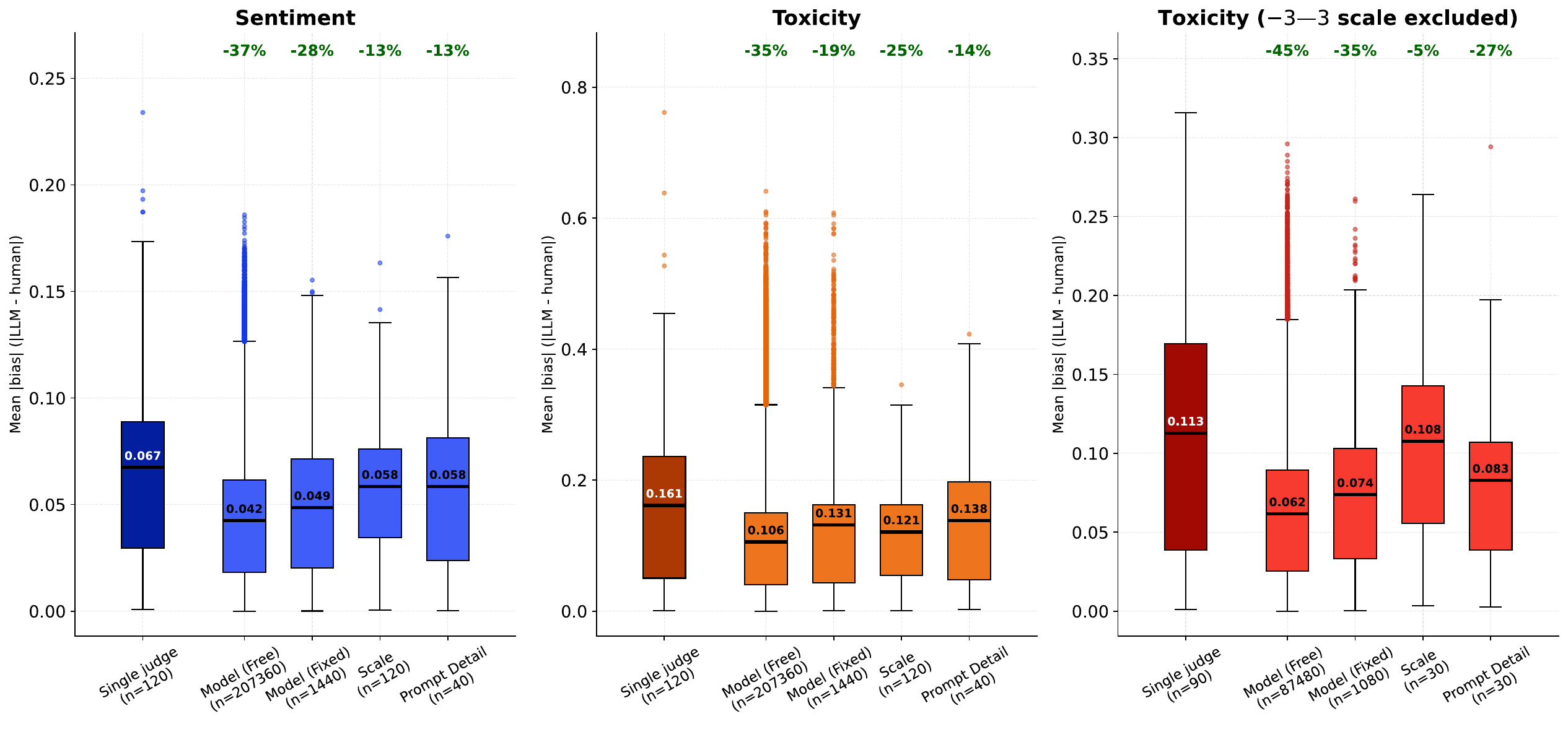}
    \end{center}
    \caption{Three-judge group results: mean absolute bias and percentage reduction relative to a single judge, for Sentiment, Toxicity, and Toxicity with the -3 -- 3 outlier scale excluded.}
    \label{fig:mitigation-boxplots}
\end{figure}

\textbf{Results.} Figure~\ref{fig:mitigation-boxplots} shows the bias produced by each rule, for Sentiment, Toxicity, and Toxicity with -3 -- 3 outlier results removed. The strongest and most reliable reduction comes from the Model (Free) strategy, which lowers mean bias by 37.0\% for Sentiment and 34.6\% for Toxicity. The Model (Fixed) strategy still gives a large reduction (27.9\% and 18.6\%) and consistently beats the Scale and Prompt Detail strategies, confirming that model variety drives most of the benefit. While scale variety looks competitive with the Model (Fixed) strategy on the full Toxicity benchmark (25.2\%), this is caused by the known outlier behavior of the -3 -- 3 scale: once that scale is excluded, the effect drops to only 4.6\%. 

\begin{figure}[h]
    \begin{center}
    \includegraphics[width=\linewidth]{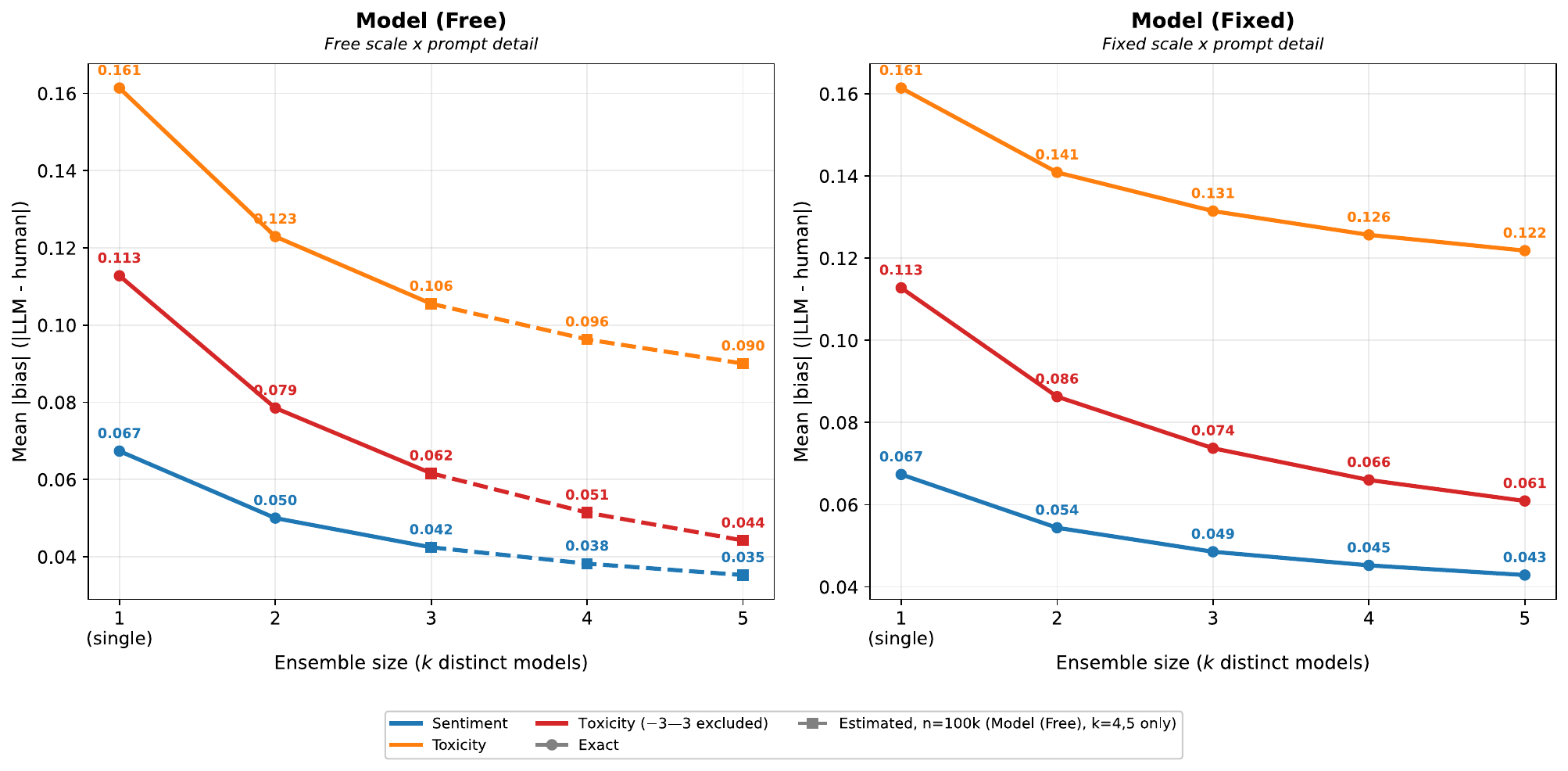}
    \end{center}
    \caption{Mean absolute bias as a function of judge-ensemble size (2 to 5 members), under the Model (Free) and Model (Fixed) strategies, for Sentiment and Toxicity.}
    \label{fig:ensemble-size-bias}
\end{figure}

\textbf{Judge ensemble size.} We estimate the influence of ensemble size by measuring the bias improvement when using 2 to 5 judges per ensemble, using the Model strategies exclusively. Due to the large number of possible 4-judge and 5-judge ensembles in the Model (Free) strategy, we estimate the result from 100,000 randomly drawn ensembles for each ensemble size rather than including every single possible ensemble. 
Under both strategies, while bias steadily decreases as the size of the ensemble increases, the gains decrease each time (see Figure~\ref{fig:ensemble-size-bias}); the move from one to two members gives the largest single improvement in every case. The Model (Free) strategy always reaches a lower bias than the Model (Fixed) one at the same group size, showing that varying other design settings within each ensemble enhances the effectiveness of this ensemble mitigation method. In practice, adding judges beyond three brings only a small further benefit, and using two-judge ensembles is enough to see an improvement.

\clearpage
\subsection{Design Choice Impact}
\label{app:design-choice-metrics}

\begin{table*}[t]
\centering
\tiny
\setlength{\tabcolsep}{3.5pt}
\caption{Design-choice impact breakdown. \% Sig / \% RBC / \% $\Delta>$.2 as in
Table~\ref{tab:star-summary}.}
\label{tab:q2-config-level-full}
\begin{tabular}{l ccc ccc ccc ccc ccc}
\toprule
& \multicolumn{9}{c}{\textbf{Bias}} & \multicolumn{6}{c}{\textbf{Error}} \\
\cmidrule(lr){2-10} \cmidrule(lr){11-16}
& \multicolumn{3}{c}{Sig} & \multicolumn{3}{c}{RBC} & \multicolumn{3}{c}{$\Delta>$.2} & \multicolumn{3}{c}{Sig} & \multicolumn{3}{c}{$\Delta>$.2} \\
\textbf{Design choice} & S & T & \textbf{Tot} & S & T & \textbf{Tot} & S & T & \textbf{Tot} & S & T & \textbf{Tot} & S & T & \textbf{Tot} \\
\midrule
\multicolumn{16}{l}{\textit{Scale}} \\
1 -- 7 & 61\% & 83\% & \textbf{72\%} & 34\% & 59\% & \textbf{47\%} & 4\% & 24\% & \textbf{14\%} & 76\% & 83\% & \textbf{79\%} & 0\% & 11\% & \textbf{6\%} \\
-3 -- 3 & 62\% & 90\% & \textbf{76\%} & 39\% & 86\% & \textbf{62\%} & 8\% & 67\% & \textbf{37\%} & 76\% & 92\% & \textbf{84\%} & 4\% & 41\% & \textbf{23\%} \\
0 -- 1 & 64\% & 80\% & \textbf{72\%} & 31\% & 60\% & \textbf{46\%} & 3\% & 29\% & \textbf{16\%} & 77\% & 74\% & \textbf{76\%} & 3\% & 18\% & \textbf{10\%} \\
0 -- 100 & 63\% & 80\% & \textbf{72\%} & 29\% & 62\% & \textbf{46\%} & 1\% & 24\% & \textbf{13\%} & 81\% & 79\% & \textbf{80\%} & 2\% & 17\% & \textbf{9\%} \\
\textbf{Scale (overall)} & \textbf{63\%} & \textbf{83\%} & \textbf{74\%} & \textbf{33\%} & \textbf{67\%} & \textbf{50\%} & \textbf{4\%} & \textbf{35\%} & \textbf{20\%} & \textbf{77\%} & \textbf{82\%} & \textbf{80\%} & \textbf{2\%} & \textbf{21\%} & \textbf{11\%} \\
\addlinespace
\multicolumn{16}{l}{\textit{Prompt Detail}} \\
Base & 50\% & 84\% & \textbf{67\%} & 21\% & 62\% & \textbf{42\%} & 0\% & 24\% & \textbf{12\%} & 49\% & 80\% & \textbf{64\%} & 0\% & 19\% & \textbf{9\%} \\
Short In-Context & 57\% & 80\% & \textbf{69\%} & 38\% & 60\% & \textbf{49\%} & 1\% & 12\% & \textbf{7\%} & 41\% & 82\% & \textbf{62\%} & 0\% & 11\% & \textbf{6\%} \\
Long In-Context & 57\% & 81\% & \textbf{69\%} & 29\% & 60\% & \textbf{44\%} & 1\% & 26\% & \textbf{14\%} & 42\% & 92\% & \textbf{68\%} & 0\% & 30\% & \textbf{15\%} \\
\textbf{Prompt Detail (overall)} & \textbf{55\%} & \textbf{82\%} & \textbf{70\%} & \textbf{29\%} & \textbf{61\%} & \textbf{45\%} & \textbf{1\%} & \textbf{21\%} & \textbf{11\%} & \textbf{44\%} & \textbf{85\%} & \textbf{65\%} & \textbf{0\%} & \textbf{20\%} & \textbf{10\%} \\
\addlinespace
\multicolumn{16}{l}{\textit{Model}} \\
Claude Haiku 4.5 & 68\% & 90\% & \textbf{79\%} & 34\% & 77\% & \textbf{56\%} & 2\% & 30\% & \textbf{16\%} & 65\% & 62\% & \textbf{63\%} & 1\% & 6\% & \textbf{4\%} \\
Claude Sonnet 5 & 84\% & 78\% & \textbf{81\%} & 62\% & 57\% & \textbf{60\%} & 2\% & 24\% & \textbf{13\%} & 54\% & 67\% & \textbf{60\%} & 0\% & 3\% & \textbf{1\%} \\
DeepSeek V4 Pro & 69\% & 79\% & \textbf{74\%} & 42\% & 67\% & \textbf{54\%} & 4\% & 32\% & \textbf{18\%} & 59\% & 85\% & \textbf{72\%} & 0\% & 15\% & \textbf{7\%} \\
Gemini 3.7 Flash & 70\% & 76\% & \textbf{73\%} & 44\% & 63\% & \textbf{53\%} & 3\% & 21\% & \textbf{12\%} & 63\% & 54\% & \textbf{58\%} & 0\% & 2\% & \textbf{1\%} \\
GLM-5.3 & 74\% & 76\% & \textbf{75\%} & 60\% & 69\% & \textbf{64\%} & 6\% & 24\% & \textbf{15\%} & 49\% & 66\% & \textbf{57\%} & 1\% & 5\% & \textbf{3\%} \\
GPT-5.5 & 70\% & 79\% & \textbf{75\%} & 47\% & 63\% & \textbf{55\%} & 6\% & 20\% & \textbf{13\%} & 60\% & 51\% & \textbf{56\%} & 0\% & 0\% & \textbf{0\%} \\
GPT-5.6 Terra & 69\% & 73\% & \textbf{71\%} & 34\% & 60\% & \textbf{47\%} & 2\% & 23\% & \textbf{12\%} & 65\% & 56\% & \textbf{60\%} & 4\% & 1\% & \textbf{2\%} \\
gpt-oss 120b & 64\% & 87\% & \textbf{75\%} & 33\% & 74\% & \textbf{54\%} & 4\% & 41\% & \textbf{22\%} & 47\% & 62\% & \textbf{55\%} & 0\% & 7\% & \textbf{4\%} \\
Kimi K3 & 61\% & 76\% & \textbf{69\%} & 39\% & 60\% & \textbf{50\%} & 5\% & 16\% & \textbf{10\%} & 46\% & 53\% & \textbf{50\%} & 1\% & 2\% & \textbf{1\%} \\
Qwen 3.8 27B & 77\% & 93\% & \textbf{85\%} & 60\% & 84\% & \textbf{72\%} & 18\% & 59\% & \textbf{38\%} & 56\% & 58\% & \textbf{57\%} & 1\% & 4\% & \textbf{2\%} \\
\textbf{Model (overall)} & \textbf{71\%} & \textbf{81\%} & \textbf{75\%} & \textbf{46\%} & \textbf{67\%} & \textbf{56\%} & \textbf{5\%} & \textbf{29\%} & \textbf{17\%} & \textbf{56\%} & \textbf{61\%} & \textbf{59\%} & \textbf{1\%} & \textbf{4\%} & \textbf{3\%} \\
\midrule
\textbf{Total} & \textbf{63\%} & \textbf{82\%} & \textbf{73\%} & \textbf{36\%} & \textbf{65\%} & \textbf{50\%} & \textbf{3\%} & \textbf{28\%} & \textbf{16\%} & \textbf{59\%} & \textbf{76\%} & \textbf{68\%} & \textbf{1\%} & \textbf{15\%} & \textbf{8\%} \\
\bottomrule
\end{tabular}
\end{table*}

We report two complementary views of design-choice impact beyond the summary in Table~\ref{tab:star-summary}. In Table~\ref{tab:q2-config-level-full}, we report how often judges differ significantly from other judges with the same design choices except one, and how often that difference reaches a meaningful effect size. This matters because a design choice can look impactful while being concentrated in a handful of settings only. Table~\ref{tab:q2-nothing-fixed} extends this from single design choices to the full space of judge designs: rather than holding every other setting fixed and varying only one
choice at a time, it compares every pair among all 120 designs regardless of how many settings differ simultaneously (Table \ref{tab:q2-nothing-fixed}). Reporting both the average and maximum pairwise difference at each level of constraint answers two questions: the average captures how much a single design choice typically shifts scores, while the maximum captures the extent to which a single choice, or an unconstrained combination of choices, can skew results in the most extreme case observed.

\begin{table}[t]
\centering
\small
\setlength{\tabcolsep}{4pt}
\caption{Disagreement magnitude across designs. Values are average and
maximum absolute pairwise difference on the 1--7 scale; $n$ is the number of
pairwise comparisons for each average.}
\label{tab:q2-nothing-fixed}
\begin{tabular}{l ccc}
\toprule
& Avg & Max & $n$ \\
\midrule
\multicolumn{4}{l}{\textit{Scale (model+family fixed)}} \\
\quad Sentiment (Bias) & 0.06 & 0.24 & 180 \\
\quad Sentiment (Error) & 0.07 & 0.28 & 180 \\
\quad Toxicity (Bias) & 0.19 & 0.93 & 180 \\
\quad Toxicity (Error) & 0.11 & 0.42 & 180 \\
\multicolumn{4}{l}{\textit{Prompt Detail (model+scale fixed)}} \\
\quad Sentiment (Bias) & 0.05 & 0.26 & 120 \\
\quad Sentiment (Error) & 0.04 & 0.19 & 120 \\
\quad Toxicity (Bias) & 0.15 & 0.71 & 120 \\
\quad Toxicity (Error) & 0.12 & 0.45 & 120 \\
\multicolumn{4}{l}{\textit{Model (family+scale fixed)}} \\
\quad Sentiment (Bias) & 0.08 & 0.39 & 540 \\
\quad Sentiment (Error) & 0.05 & 0.26 & 540 \\
\quad Toxicity (Bias) & 0.15 & 0.53 & 540 \\
\quad Toxicity (Error) & 0.07 & 0.28 & 540 \\
\addlinespace
\multicolumn{4}{l}{\textit{Nothing fixed -- any two of all 120 designs}} \\
\quad \textbf{Sentiment} (\textbf{Bias}) & \textbf{0.09} & \textbf{0.43} & \textbf{7140} \\
\quad \textbf{Sentiment} (\textbf{Error}) & \textbf{0.07} & \textbf{0.36} & \textbf{7140} \\
\quad \textbf{Toxicity} (\textbf{Bias}) & \textbf{0.22} & \textbf{1.08} & \textbf{7140} \\
\quad \textbf{Toxicity} (\textbf{Error}) & \textbf{0.14} & \textbf{0.56} & \textbf{7140} \\
\bottomrule
\end{tabular}
\end{table}

\clearpage
\section{Accuracy Classification Task: Extended Results}
\label{app:reasoning_results}
\subsection{LLM Judge Performance}
\label{app:reasoning_results_performance}

\begin{table*}[!ht]
\centering
\footnotesize
\setlength{\tabcolsep}{2pt}
\caption{Accuracy, leniency, and error count on the accuracy classification task, split by benchmark subset, with significance stars where accuracy is below the column's accuracy (significance representation and test described in Section~\ref{sec:methods}).}
\label{tab:reasoning-accuracy-leniency-full}
\begin{tabular}{l ccc ccc ccc}
\toprule
& \multicolumn{3}{c}{Total} & \multicolumn{3}{c}{Nearly Agree} & \multicolumn{3}{c}{Ambiguous} \\
\cmidrule(lr){2-4} \cmidrule(lr){5-7} \cmidrule(lr){8-10}
& Acc. & Len. & \#Err & Acc. & Len. & \#Err & Acc. & Len. & \#Err \\
\midrule
\multicolumn{10}{l}{\textit{Prompt Detail}} \\
Base              & 96.6\% & 81.3\% & 406 & 98.1\% & 70.2\% & 47 & 61.8\% & 82.7\% & 359 \\
Short In-Context  & 96.8\% & 73.3\% & 386 & 97.7\% & 42.1\% & 57 & 65.0\% & 78.7\% & 329 \\
Long In-Context   & 96.0\% & 52.4\% & 475 & 93.6\% & 12.6\% & 159 & 66.3\% & 72.5\% & 316 \\
\midrule
\multicolumn{10}{l}{\textit{Reasoning Effort}} \\
Default  & 96.5\% & 69.2\% & 623 & 96.8\% & 29.4\% & 119 & 64.2\% & 78.6\% & 504 \\
Low               & 96.4\% & 66.9\% & 644 & 96.1\%$^{*}$ & 29.2\% & 144 & 64.5\% & 77.8\% & 500 \\
\midrule
\multicolumn{10}{l}{\textit{Model}} \\
GLM-5.3, Default & 96.7\% & 86.4\% & 59 & 99.2\% & 33.3\% & 3  & 60.3\% & 89.3\% & 56 \\
GLM-5.3, Low  & 96.6\% & 87.1\% & 62 & 96.8\% & 83.3\% & 12 & 64.5\% & 88.0\% & 50 \\
\rowcolor{gray!12}
\textbf{GLM-5.3} & \textbf{96.7\%} & \textbf{86.8\%} & \textbf{121} & \textbf{98.0\%} & \textbf{73.3\%} & \textbf{15} & \textbf{62.3\%} & \textbf{88.7\%} & \textbf{106} \\
\addlinespace
Claude Haiku 4.5, Default & 96.5\% & 54.0\% & 63 & 94.4\%$^{*}$ & 52.4\% & 21 & 70.2\% & 54.8\% & 42 \\
Claude Haiku 4.5, Low  & 96.1\% & 62.9\% & 70 & 94.9\% & 57.9\% & 19 & 63.8\% & 64.7\% & 51 \\
\rowcolor{gray!12}
\textbf{Claude Haiku 4.5} & \textbf{96.3\%} & \textbf{58.6\%} & \textbf{133} & \textbf{94.6\%$^{**}$} & \textbf{55.0\%} & \textbf{40} & \textbf{67.4\%} & \textbf{60.2\%} & \textbf{93} \\
\addlinespace
Claude Sonnet 5, Default & 96.2\% & 82.4\% & 68 & 97.8\% & 37.5\% & 8  & 57.1\% & 88.3\% & 60 \\
Claude Sonnet 5, Low  & 96.4\% & 75.4\% & 65 & 96.5\% & 15.4\% & 13 & 63.1\% & 90.4\% & 52 \\
\rowcolor{gray!12}
\textbf{Claude Sonnet 5} & \textbf{96.3\%} & \textbf{78.9\%} & \textbf{133} & \textbf{97.2\%} & \textbf{23.8\%} & \textbf{21} & \textbf{59.8\%} & \textbf{89.3\%} & \textbf{112} \\
\addlinespace
DeepSeek V4 Pro, Default & 96.1\% & 38.6\% & 70 & 92.2\%$^{***}$ & 6.9\%  & 29 & 70.9\% & 61.0\% & 41 \\
DeepSeek V4 Pro, Low  & 96.7\% & 50.8\% & 59 & 93.8\%$^{*}$ & 26.1\% & 23 & 74.5\% & 66.7\% & 36 \\
\rowcolor{gray!12}
\textbf{DeepSeek V4 Pro} & \textbf{96.4\%} & \textbf{44.2\%} & \textbf{129} & \textbf{93.0\%$^{***}$} & \textbf{15.4\%} & \textbf{52} & \textbf{72.5\%} & \textbf{63.6\%} & \textbf{77} \\
\addlinespace
Gemini 3.7 Flash, Default & 93.9\%$^{***}$ & 97.2\% & 109 & 95.7\% & 100.0\% & 16 & 34.0\%$^{***}$ & 96.8\% & 93 \\
Gemini 3.7 Flash, Low  & 94.2\%$^{***}$ & 97.1\% & 105 & 97.0\% & 100.0\% & 11 & 33.3\%$^{***}$ & 96.8\% & 94 \\
\rowcolor{gray!12}
\textbf{Gemini 3.7 Flash} & \textbf{94.1\%$^{***}$} & \textbf{97.2\%} & \textbf{214} & \textbf{96.4\%} & \textbf{100.0\%} & \textbf{27} & \textbf{33.7\%$^{***}$} & \textbf{96.8\%} & \textbf{187} \\
\addlinespace
GPT-5.5, Default & 97.1\% & 71.7\% & 53 & 99.7\% & 0.0\% & 1 & 63.1\% & 73.1\% & 52 \\
GPT-5.5, Low  & 96.9\% & 67.3\% & 55 & 98.9\% & 0.0\% & 4 & 63.8\% & 72.5\% & 51 \\
\rowcolor{gray!12}
\textbf{GPT-5.5} & \textbf{97.0\%} & \textbf{69.4\%} & \textbf{108} & \textbf{99.3\%} & \textbf{0.0\%} & \textbf{5} & \textbf{63.5\%} & \textbf{72.8\%} & \textbf{103} \\
\addlinespace
GPT-5.6 Terra, Default & 96.1\% & 44.3\% & 70 & 94.1\%$^{*}$ & 0.0\% & 22 & 66.0\% & 64.6\% & 48 \\
GPT-5.6 Terra, Low  & 95.1\%$^{**}$ & 38.6\% & 88 & 90.6\%$^{***}$ & 0.0\% & 35 & 62.4\% & 64.2\% & 53 \\
\rowcolor{gray!12}
\textbf{GPT-5.6 Terra} & \textbf{95.6\%$^{*}$} & \textbf{41.1\%} & \textbf{158} & \textbf{92.3\%$^{***}$} & \textbf{0.0\%} & \textbf{57} & \textbf{64.3\%} & \textbf{64.4\%} & \textbf{101} \\
\addlinespace
gpt-oss 120b, Default & 98.0\% & 38.9\% & 36 & 96.8\% & 0.0\% & 12 & 83.0\% & 58.3\% & 24 \\
gpt-oss 120b, Low  & 97.9\% & 47.4\% & 38 & 97.6\% & 0.0\% & 9  & 79.4\% & 62.1\% & 29 \\
\rowcolor{gray!12}
\textbf{gpt-oss 120b} & \textbf{97.9\%} & \textbf{43.2\%} & \textbf{74} & \textbf{97.2\%} & \textbf{0.0\%} & \textbf{21} & \textbf{81.4\%} & \textbf{60.4\%} & \textbf{53} \\
\addlinespace
Kimi K3, Default & 97.4\% & 78.3\% & 46 & 99.5\% & 50.0\% & 2 & 68.8\% & 79.5\% & 44 \\
Kimi K3, Low  & 97.4\% & 76.1\% & 46 & 99.5\% & 0.0\%  & 2 & 68.8\% & 79.5\% & 44 \\
\rowcolor{gray!12}
\textbf{Kimi K3} & \textbf{97.4\%} & \textbf{77.2\%} & \textbf{92} & \textbf{99.5\%} & \textbf{25.0\%} & \textbf{4} & \textbf{68.8\%} & \textbf{79.5\%} & \textbf{88} \\
\addlinespace
Qwen 3.8 27B, Default & 97.3\% & 77.6\% & 49 & 98.7\% & 20.0\% & 5  & 68.8\% & 84.1\% & 44 \\
Qwen 3.8 27B, Low  & 96.9\% & 50.0\% & 56 & 95.7\% & 12.5\% & 16 & 71.6\% & 65.0\% & 40 \\
\rowcolor{gray!12}
\textbf{Qwen 3.8 27B} & \textbf{97.1\%} & \textbf{62.9\%} & \textbf{105} & \textbf{97.2\%} & \textbf{14.3\%} & \textbf{21} & \textbf{70.2\%} & \textbf{75.0\%} & \textbf{84} \\
\addlinespace
\midrule
\textbf{Total} & \textbf{96.5\%} & \textbf{68.0\%} & \textbf{1267} & \textbf{96.5\%} & \textbf{29.3\%} & \textbf{263} & \textbf{64.4\%} & \textbf{78.2\%} & \textbf{1004} \\
\bottomrule
\end{tabular}
\end{table*}

\begin{table*}[t]
\centering
\small
\caption{Accuracy, leniency, error count, and directional switch rates per model (reasoning-effort levels aggregated) for the full benchmark. False$\to$True and True$\to$False error switch rates for In-Context prompts are relative to the base prompt.}
\label{tab:reasoning-per-model-total}
\begin{tabular}{@{}l ccccc@{}}
\toprule
Model & Accuracy & Leniency & Error Count & False$\to$True & True$\to$False \\
\midrule
\multicolumn{6}{l}{\textit{Claude Haiku 4.5}} \\
\quad Base & 96.0\% & 66.7\% & 48 & --- & --- \\
\quad Short & 96.6\% & 68.3\% & 41 & 1.3\% & 1.4\% \\
\quad Long & 96.3\% & 40.9\% & 44 & 0.5\% & 2.5\% \\
\multicolumn{6}{l}{\textit{Claude Sonnet 5}} \\
\quad Base & 96.2\% & 73.9\% & 46 & --- & --- \\
\quad Short & 96.1\% & 87.2\% & 47 & 1.5\% & 0.4\% \\
\quad Long & 96.7\% & 75.0\% & 40 & 1.1\% & 1.3\% \\
\multicolumn{6}{l}{\textit{DeepSeek V4 Pro}} \\
\quad Base & 98.1\% & 82.6\% & 23 & --- & --- \\
\quad Short & 97.5\% & 53.3\% & 30 & 0.5\% & 1.6\% \\
\quad Long & 93.7\% & 28.9\% & 76 & 1.0\% & 4.9\% \\
\multicolumn{6}{l}{\textit{Gemini 3.7 Flash}} \\
\quad Base & 93.7\% & 97.4\% & 76 & --- & --- \\
\quad Short & 93.7\% & 97.3\% & 75 & 0.7\% & 0.8\% \\
\quad Long & 94.7\% & 96.8\% & 63 & 0.5\% & 1.6\% \\
\multicolumn{6}{l}{\textit{GLM-5.3}} \\
\quad Base & 96.0\% & 89.6\% & 48 & --- & --- \\
\quad Short & 97.4\% & 83.9\% & 31 & 0.5\% & 1.9\% \\
\quad Long & 96.5\% & 85.7\% & 42 & 0.9\% & 1.6\% \\
\multicolumn{6}{l}{\textit{GPT-5.5}} \\
\quad Base & 96.8\% & 81.6\% & 38 & --- & --- \\
\quad Short & 97.2\% & 76.5\% & 34 & 0.2\% & 0.7\% \\
\quad Long & 97.0\% & 50.0\% & 36 & 0.0\% & 2.0\% \\
\multicolumn{6}{l}{\textit{GPT-5.6 Terra}} \\
\quad Base & 97.4\% & 71.0\% & 31 & --- & --- \\
\quad Short & 95.9\% & 42.9\% & 49 & 0.5\% & 2.2\% \\
\quad Long & 93.5\% & 28.2\% & 78 & 0.8\% & 4.8\% \\
\multicolumn{6}{l}{\textit{gpt-oss 120b}} \\
\quad Base & 97.9\% & 64.0\% & 25 & --- & --- \\
\quad Short & 98.3\% & 60.0\% & 20 & 0.6\% & 0.8\% \\
\quad Long & 97.6\% & 13.8\% & 29 & 0.3\% & 2.6\% \\
\multicolumn{6}{l}{\textit{Kimi K3}} \\
\quad Base & 97.1\% & 82.9\% & 35 & --- & --- \\
\quad Short & 97.7\% & 75.0\% & 28 & 0.4\% & 1.2\% \\
\quad Long & 97.6\% & 72.4\% & 29 & 0.3\% & 1.1\% \\
\multicolumn{6}{l}{\textit{Qwen 3.8 27B}} \\
\quad Base & 97.0\% & 83.3\% & 36 & --- & --- \\
\quad Short & 97.4\% & 61.3\% & 31 & 0.7\% & 2.1\% \\
\quad Long & 96.8\% & 44.7\% & 38 & 0.3\% & 2.7\% \\
\midrule
\multicolumn{6}{l}{\textbf{Total}} \\
\quad \textbf{Base} & 96.6\% & 81.3\% & 406 & --- & --- \\
\quad \textbf{Short} & 96.8\% & 73.3\% & 386 & 0.7\% & 1.3\% \\
\quad \textbf{Long} & 96.0\% & 52.4\% & 475 & 0.6\% & 2.5\% \\
\bottomrule
\end{tabular}
\end{table*}

\begin{table*}[t]
\centering
\small
\caption{Accuracy, leniency, error count, and directional switch rates per model (reasoning-effort levels aggregated) for the near-agreement benchmark subset. False$\to$True and True$\to$False error switch rates for In-Context prompts are relative to the base prompt.}
\label{tab:reasoning-per-model-near}
\begin{tabular}{@{}l ccccc@{}}
\toprule
Model & Accuracy & Leniency & Error Count & False$\to$True & True$\to$False \\
\midrule
\multicolumn{6}{l}{\textit{Claude Haiku 4.5}} \\
\quad Base & 94.8\% & 69.2\% & 13 & --- & --- \\
\quad Short & 96.0\% & 80.0\% & 10 & 4.0\% & 3.6\% \\
\quad Long & 93.1\% & 29.4\% & 17 & 1.2\% & 6.1\% \\
\multicolumn{6}{l}{\textit{Claude Sonnet 5}} \\
\quad Base & 96.8\% & 12.5\% & 8 & --- & --- \\
\quad Short & 97.6\% & 50.0\% & 6 & 3.2\% & 0.8\% \\
\quad Long & 97.2\% & 14.3\% & 7 & 2.0\% & 1.6\% \\
\multicolumn{6}{l}{\textit{DeepSeek V4 Pro}} \\
\quad Base & 99.6\% & 100.0\% & 1 & --- & --- \\
\quad Short & 97.2\% & 28.6\% & 7 & 0.8\% & 2.4\% \\
\quad Long & 82.3\% & 11.4\% & 44 & 2.0\% & 16.1\% \\
\multicolumn{6}{l}{\textit{Gemini 3.7 Flash}} \\
\quad Base & 94.8\% & 100.0\% & 13 & --- & --- \\
\quad Short & 96.4\% & 100.0\% & 9 & 1.2\% & 2.8\% \\
\quad Long & 98.0\% & 100.0\% & 5 & 0.8\% & 4.0\% \\
\multicolumn{6}{l}{\textit{GLM-5.3}} \\
\quad Base & 97.2\% & 100.0\% & 7 & --- & --- \\
\quad Short & 100.0\% & --- & 0 & 0.0\% & 2.8\% \\
\quad Long & 96.8\% & 50.0\% & 8 & 0.8\% & 3.6\% \\
\multicolumn{6}{l}{\textit{GPT-5.5}} \\
\quad Base & 100.0\% & --- & 0 & --- & --- \\
\quad Short & 100.0\% & --- & 0 & 0.0\% & 0.0\% \\
\quad Long & 98.0\% & 0.0\% & 5 & 0.0\% & 2.0\% \\
\multicolumn{6}{l}{\textit{GPT-5.6 Terra}} \\
\quad Base & 99.6\% & 0.0\% & 1 & --- & --- \\
\quad Short & 94.0\% & 0.0\% & 15 & 0.4\% & 6.0\% \\
\quad Long & 83.5\% & 0.0\% & 41 & 0.4\% & 16.5\% \\
\multicolumn{6}{l}{\textit{gpt-oss 120b}} \\
\quad Base & 99.2\% & 0.0\% & 2 & --- & --- \\
\quad Short & 99.2\% & 0.0\% & 2 & 0.4\% & 0.4\% \\
\quad Long & 93.1\% & 0.0\% & 17 & 0.4\% & 6.5\% \\
\multicolumn{6}{l}{\textit{Kimi K3}} \\
\quad Base & 99.6\% & 100.0\% & 1 & --- & --- \\
\quad Short & 99.6\% & 0.0\% & 1 & 0.0\% & 0.8\% \\
\quad Long & 99.2\% & 0.0\% & 2 & 0.0\% & 1.2\% \\
\multicolumn{6}{l}{\textit{Qwen 3.8 27B}} \\
\quad Base & 99.6\% & 100.0\% & 1 & --- & --- \\
\quad Short & 97.2\% & 28.6\% & 7 & 0.8\% & 2.4\% \\
\quad Long & 94.8\% & 0.0\% & 13 & 0.0\% & 5.6\% \\
\midrule
\multicolumn{6}{l}{\textbf{Total}} \\
\quad \textbf{Base} & 98.1\% & 70.2\% & 47 & --- & --- \\
\quad \textbf{Short} & 97.7\% & 42.1\% & 57 & 1.1\% & 2.2\% \\
\quad \textbf{Long} & 93.6\% & 12.6\% & 159 & 0.8\% & 6.3\% \\
\bottomrule
\end{tabular}
\end{table*}

\begin{table*}[t]
\centering
\small
\caption{Accuracy, leniency, error count, and directional switch rates per model (reasoning-effort levels aggregated) for the ambiguous benchmark subset. False$\to$True and True$\to$False error switch rates for In-Context prompts are relative to the base prompt.}
\label{tab:reasoning-per-model-ambiguous}
\begin{tabular}{@{}l ccccc@{}}
\toprule
Model & Accuracy & Leniency & Error Count & False$\to$True & True$\to$False \\
\midrule
\multicolumn{6}{l}{\textit{Claude Haiku 4.5}} \\
\quad Base & 62.8\% & 65.7\% & 35 & --- & --- \\
\quad Short & 67.0\% & 64.5\% & 31 & 6.4\% & 8.5\% \\
\quad Long & 71.3\% & 48.1\% & 27 & 3.2\% & 16.0\% \\
\multicolumn{6}{l}{\textit{Claude Sonnet 5}} \\
\quad Base & 59.6\% & 86.8\% & 38 & --- & --- \\
\quad Short & 56.4\% & 92.7\% & 41 & 10.6\% & 3.2\% \\
\quad Long & 64.5\% & 87.9\% & 33 & 8.6\% & 11.8\% \\
\multicolumn{6}{l}{\textit{DeepSeek V4 Pro}} \\
\quad Base & 76.6\% & 81.8\% & 22 & --- & --- \\
\quad Short & 75.5\% & 60.9\% & 23 & 4.3\% & 13.8\% \\
\quad Long & 66.0\% & 53.1\% & 32 & 7.4\% & 20.2\% \\
\multicolumn{6}{l}{\textit{Gemini 3.7 Flash}} \\
\quad Base & 33.0\% & 96.8\% & 63 & --- & --- \\
\quad Short & 29.8\% & 97.0\% & 66 & 5.3\% & 2.1\% \\
\quad Long & 38.3\% & 96.6\% & 58 & 4.3\% & 9.6\% \\
\multicolumn{6}{l}{\textit{GLM-5.3}} \\
\quad Base & 56.4\% & 87.8\% & 41 & --- & --- \\
\quad Short & 67.0\% & 83.9\% & 31 & 6.4\% & 17.0\% \\
\quad Long & 63.8\% & 94.1\% & 34 & 9.6\% & 10.6\% \\
\multicolumn{6}{l}{\textit{GPT-5.5}} \\
\quad Base & 59.6\% & 81.6\% & 38 & --- & --- \\
\quad Short & 63.8\% & 76.5\% & 34 & 2.1\% & 8.5\% \\
\quad Long & 67.0\% & 58.1\% & 31 & 0.0\% & 20.2\% \\
\multicolumn{6}{l}{\textit{GPT-5.6 Terra}} \\
\quad Base & 68.1\% & 73.3\% & 30 & --- & --- \\
\quad Short & 63.8\% & 61.8\% & 34 & 5.3\% & 11.7\% \\
\quad Long & 60.6\% & 59.5\% & 37 & 9.6\% & 17.0\% \\
\multicolumn{6}{l}{\textit{gpt-oss 120b}} \\
\quad Base & 75.5\% & 69.6\% & 23 & --- & --- \\
\quad Short & 80.9\% & 66.7\% & 18 & 6.4\% & 9.6\% \\
\quad Long & 87.2\% & 33.3\% & 12 & 2.1\% & 16.0\% \\
\multicolumn{6}{l}{\textit{Kimi K3}} \\
\quad Base & 63.8\% & 82.4\% & 34 & --- & --- \\
\quad Short & 71.3\% & 77.8\% & 27 & 5.3\% & 12.8\% \\
\quad Long & 71.3\% & 77.8\% & 27 & 3.2\% & 10.6\% \\
\multicolumn{6}{l}{\textit{Qwen 3.8 27B}} \\
\quad Base & 62.8\% & 82.9\% & 35 & --- & --- \\
\quad Short & 74.5\% & 70.8\% & 24 & 6.4\% & 20.2\% \\
\quad Long & 73.4\% & 68.0\% & 25 & 4.3\% & 19.1\% \\
\midrule
\multicolumn{6}{l}{\textbf{Total}} \\
\quad \textbf{Base} & 61.8\% & 82.7\% & 359 & --- & --- \\
\quad \textbf{Short} & 65.0\% & 78.7\% & 329 & 5.9\% & 10.7\% \\
\quad \textbf{Long} & 66.3\% & 72.5\% & 316 & 5.2\% & 15.1\% \\
\bottomrule
\end{tabular}
\end{table*}

\clearpage
\subsection{Design Choice Impact}
\label{app:design-choice-classificationtask}

\begin{table*}[ht!]
\centering
\small
\setlength{\tabcolsep}{4pt}
\caption{Full breakdown of pairwise robustness scores for each design choice. \textbf{Avg.} is the mean pairwise impact across all pairs of settings. \textbf{Max.} is the single largest pairwise impact observed anywhere in the design choice. For the reasoning effort category, which has only two settings, both statistics coincide since there is only one pair.}
\label{tab:reasoning-robustness-full-breakdown}
\begin{tabular}{@{}l cc cc cc@{}}
\toprule
& \multicolumn{2}{c}{Accuracy} & \multicolumn{2}{c}{Switch Rate} & \multicolumn{2}{c}{Leniency} \\
\cmidrule(lr){2-3} \cmidrule(lr){4-5} \cmidrule(lr){6-7}
\textbf{Design choice} & Avg. & Max. & Avg. & Max. & Avg. & Max. \\
\midrule
\multicolumn{7}{l}{\textit{Total}} \\
Prompt Detail     & 0.4\% & 0.6\%  & 2.5\%  & 3.1\%  & 18.1\% & 28.9\% \\
Reasoning Effort  & 0.1\% & 0.1\%  & 2.0\%  & 2.0\%  & 2.3\%  & 2.3\%  \\
Model             & 1.2\% & 3.9\%  & 4.1\%  & 7.1\%  & 23.2\% & 56.1\% \\
\midrule
\multicolumn{7}{l}{\textit{Nearly Agree}} \\
Prompt Detail     & 2.7\% & 4.0\%  & 5.4\%  & 6.6\%  & 43.3\% & 65.0\% \\
Reasoning Effort  & 0.7\% & 0.7\%  & 4.8\%  & 4.8\%  & 0.2\%  & 0.2\%  \\
Model             & 2.8\% & 7.1\%  & 6.2\%  & 12.9\% & 38.4\% & 100.0\% \\
\midrule
\multicolumn{7}{l}{\textit{Ambiguous}} \\
Prompt Detail     & 2.1\% & 3.1\%  & 17.5\% & 21.5\% & 5.6\%  & 8.5\%  \\
Reasoning Effort  & 0.3\% & 0.3\%  & 13.5\% & 13.5\% & 0.8\%  & 0.8\%  \\
Model             & 12.7\% & 47.5\% & 36.5\% & 66.0\% & 15.7\% & 36.6\% \\
\bottomrule
\end{tabular}
\end{table*}

\end{document}